\pdfoutput=1
\documentclass[11pt]{article}

\usepackage{acl}

\usepackage{times}
\usepackage{latexsym}

\usepackage[T1]{fontenc}
\usepackage[utf8]{inputenc}

\usepackage{microtype}

\usepackage{inconsolata}

\usepackage{graphicx}

\usepackage{times}
\usepackage{latexsym}
\usepackage{enumitem}
\usepackage{multirow}
\usepackage{array}
\usepackage{amsfonts}
\usepackage{subcaption}
\usepackage{makecell}
\usepackage{soul}
\usepackage{xcolor,colortbl}
\usepackage{xspace}
\usepackage{textcomp}
\usepackage{stfloats}
\usepackage{url}
\usepackage{verbatim}
\usepackage{graphicx}
\usepackage{algorithm}
\usepackage{algorithmic}
\usepackage{amssymb}
\usepackage{helvet}  % DO NOT CHANGE THIS
\usepackage{courier}  % DO NOT CHANGE THIS
\usepackage{newfloat}
\usepackage{listings}
\usepackage{multirow}
\usepackage{amsmath}
\usepackage{booktabs}
\usepackage{pifont}
\usepackage{color}
\usepackage{bbding}
\usepackage{listings}
\usepackage[export]{adjustbox}

\title{Prejudge-Before-Think: Enhancing Large Language Models at Test-Time by Process Prejudge Reasoning}

\author{Jianing Wang$^{1}$\dag, Jin Jiang$^{1,2}$, Yang Liu$^{1}$, Mengdi Zhang$^{1}$, Xunliang Cai$^{1}$\\
  $^1$ Meituan, Beijing, China \\
  $^2$ Peking University, Beijing, China \\
  \texttt{lygwjn@gmail.com, jiangjin@stu.pku.edu.cn} \\
  \texttt{\{liuyang509, zhangmengdi02, caixunliang\}@meituan.com}
}

\begin{document}
\maketitle
\begin{abstract}
% 在复杂推理中，我们引入一个过程预判的概念，并展现出大模型在推理过程中适当的进行预判可以显著地提升模型推理能力。
% Bootstrapping has recently been attracted by the natural language community, which has become the cornerstone of data synthesis and sampling procedures in pos-training.
% The step-by-step reasoning has been widely used to solve complex reasoning tasks in large language models (LLMs),

% The AI community has witnessed the evolution of large language models (LLMs) on complex tasks by using step-by-step reasoning. 
In this paper, we introduce a new \emph{process prejudge} strategy in LLM reasoning to demonstrate that bootstrapping with process prejudge allows the LLM to adaptively anticipate the errors encountered when advancing the subsequent reasoning steps, similar to people sometimes pausing to think about what mistakes may occur and how to avoid them, rather than relying solely on trial and error. Specifically, we define a prejudge node in the rationale, which represents a reasoning step, with at least one step that follows the prejudge node that has no paths toward the correct answer. To synthesize the prejudge reasoning process, we present an automated reasoning framework with a dynamic tree-searching strategy. This framework requires only one LLM to perform answer judging, response critiquing, prejudge generation, and thought completion. Furthermore, we develop a two-phase training mechanism with supervised fine-tuning (SFT) and reinforcement learning (RL) to further enhance the reasoning capabilities of LLMs. Experimental results from competition-level complex reasoning demonstrate that our method can teach the model to prejudge before thinking and significantly enhance the reasoning ability of LLMs~\footnote{Code and data is released at \url{https://github.com/wjn1996/Prejudge-Before-Think}.}.

\end{abstract}

\section{Introduction}

% 最近大模型在在AI社区内广泛的应用在了复杂推理领域中，
% Recently, large language models (LLMs) have made great success in solving complex reasoning problems by using 
Large language models (LLMs) have made great inroads in solving natural language processing (NLP) tasks~\cite{Brown2020Language, Ouyang2022Training, OpenAI2023GPT4}, but still struggle to produce accurate answers to complex reasoning problems.
Prior research tackles this challenge by designing well-crafted prompts to elicit the LLM to follow a step-by-step thinking paradigm, such as in-context learning~\cite{Liu2023Pre}, chain-of-thought~\cite{Wei2022Chain, Wang2023Self, Wang2024InstructGraph}, and agentic learning~\cite{Park2023Generative}.
However, these approaches akin to the System 1-style \emph{fast-thinking} paradigm~\cite{kahneman2011thinking} inevitably bring errors to the inherent steps, making it hard to generate accurate and complete solutions in one breath.

% on competition-level mathematics or complex logic problems

% has been key ingredients of recent revolution

% MCTS~\cite{Coulom2006Efficient}

% 在近几年里，诸多工作尝试通过提示工程、指令微调等方式来激活LLM的推理能力。发现通过推理形态上的设计可以较好地引导LLM进行思考。
% 最近一些工作从人类的system2思考慢思考的行为认知上得到启发，并探索出了一系列慢思考的策略，包括self-refine、reflexion，critic。LLM在解决推理任务的时候，适当的对推理的步骤进行反思和检查，可以有效地提高解决问题的准确性。
% 为了进一步提升，o1、qwq、deepseek r1等方法则再次基础上引入了test-time scaling和post-training，将这种慢思考的能力通过精调的方式让模型自发地探索出更多推理形态。

\begin{figure}[t]
  \includegraphics[width=\columnwidth]{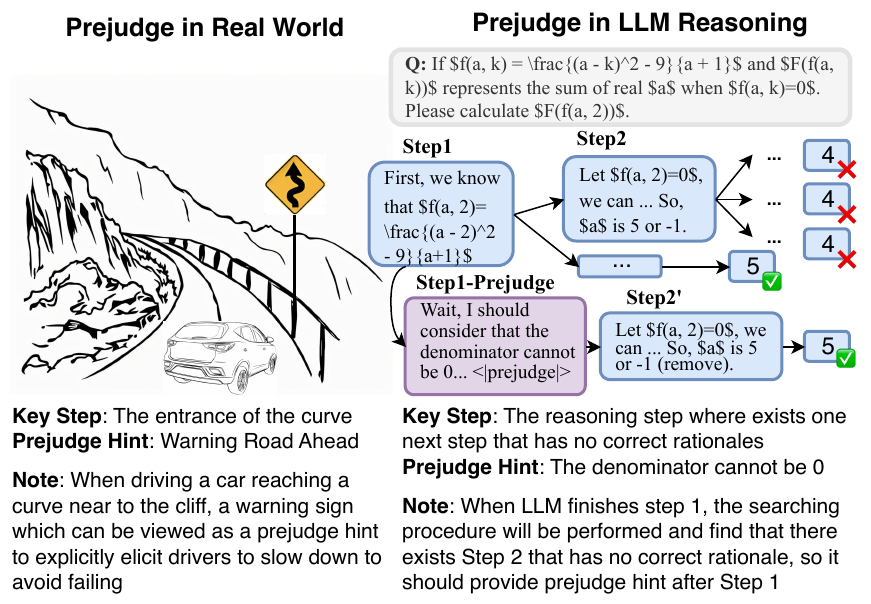}
  \caption{Examples of prejudge in the scenarios of the real world and LLM reasoning.}
  \label{fig:prejudge}
\end{figure}

% Recent research has demonstrated that the conventional reasoning methods 
Inspired by human recognition of System 2~\cite{kahneman2011thinking}, which is denoted as a \emph{slow-thinking} paradigm emulates human reasoning through a slower and deeper thought process~\cite{Zelikman2024Quiet}, most of the works have unveiled that 
% increasing the cost at test-time by 
extending the reasoning with verification, critiquing, and refinement components can significantly enhance the reasoning capability~\cite{Cobbe2021Training, Lightman2024Let, Snell2024Scaling, Qi2024Mutual, Shinn2023Reflexion, Gou2024CRITIC, Plaat2024Reasoning, Madaan2023Self}.
One major benefit is that these components can offer precise feedback, which is a valuable signal that enables the LLM to adapt or roll back the current of thought opportunely~\cite{Kumar2024Training, Wang2024Math, Chen2024Toward}.
In addition, a series of research (e.g., OpenAI's o1~\cite{Jaech2024OpenAI}, DeepSeek R1~\cite{guo2025deepseek}) has explored that post-training in the supervised fine-tuning (SFT) or reinforcement learning (RL) stage can inject these capabilities into model parameters and achieve high grades by scaling test-time cost, which spurred the development of System 2-like reasoning~\cite{Brown2024Large, Snell2024Scaling, Shao2024DeepSeekMath, Rafailov2023Direct}. 
Despite this success, the generated responses expose that LLMs tend to frequently prefer trial and error, leading to redundant error and reflection information, which is not very advocated in human consciousness.

In this paper, we introduce a new thought mode named~\emph{process prejudge}~\footnote{Notely, the ``prejudge'' in this paper means the ``predictive ability'' instead of ``prejudice'', we use the word ``prejudge'' aim to distinguish it from “predict” in machine learning.}, which is defined as prior consideration or judgment about what is about to happen in the subsequence reasoning steps.
% This pattern is inspired by human recognition that sometimes pauses to think whether it will fall into trouble and how to prevent it in subsequent reasoning steps.
In the real world, this capability aims to help people learn from past experiences and improve the accuracy of each thinking step when solving similar problems in the future. It is usually acquired after repeated interaction with the environment.
Take a vivid example illustrated in Figure~\ref{fig:prejudge}, when driving a vehicle and reaching the entrance of a curve close to a cliff, an experienced driver will slow down in advance. This is a prejudge action based on the experience that the vehicle will fall off the cliff due to inertia.
Therefore, a natural question arises:~\emph{is process prejudge useful for LLM in reasoning scenarios?} 

To reach this goal, 
% we aim at synthesizing large-scale step-by-step reasoning data with~\emph{process prejudge} and train the model with some post-training approaches to explore the performance of reasoning under prejudgment.
% To elaborate, we
we first define a~\emph{prejudge node} in the rationale, which is a specific reasoning step, and at least one step follows the prejudge node that has no path toward the correct answer.
% This means that LLM will fall into endless errors after this step, so it is necessary to provide prejudgement in advance to avoid errors.
For example in Figure~\ref{fig:prejudge}, the LLM may make mistakes at ``Step 2'' and it can be prevented when prompted with a prejudge hint as ``The denominator cannot be 0''.
To synthesize large-scale step-by-step reasoning data with~\emph{process prejudge}, we then propose an automatic reasoning framework with a dynamic tree-searching strategy, which is similar to Monto Carlo Tree Search (MCTS)~\cite{Kocsis2006Bandit, Silver2016Mastering} but needs only one LLM to perform thinking, critiquing, prejudging and verifying during searching.

Ultimately, we construct 234k data from multiple open-source datasets to train the LLM with SFT and RL techniques. The extensive experiments conducted on mathematics and logic reasoning demonstrate that the paradigm of \emph{prejudge before use} can substantially boost the LLM's reasoning ability.

\section{Preliminary}

\subsection{LLM Reasoning with Textual Rationale}

Given a LLM $\pi_{\theta}(\cdot)$ which is a transformer-based pre-trained model to map the input prompt to generated text, where $\theta$ is the parameters. 
For the reasoning task, given a question $\mathcal{Q}$, the LLM can provide a step-by-step reasoning chain consisting of $T$ intermediate step, and the entire chain can be formed as $\mathcal{Z}=[z_1, \cdots, z_T]$, where $z_{i}$ ($i\in[1, T]$) means the specific step\footnote{To elicit the LLM to generate this chain, the question $\mathcal{Q}$ should be articulated through a well-crafted instruction prompt or template. We omit this component to minimize the use of variables in writing.}.
The reasoning chain can be step-by-step generated by the LLM as $z_{i}=\pi_{\theta}(\mathcal{I}=\mathbb{I}(\mathcal{Q}, z_1, \cdots, z_{i-1}))$, where $\mathbb{I}(\cdots)$ is function to concate all generated prefix sequences to form the input prompt $\mathcal{I}$.

\subsection{Bootstrapping with Tree Searching}

% Tree searching is a bootstrapping method for mining the candidate's rationale for the answer.
Suppose that the prefix reasoning steps of $\mathcal{Q}$ are $\mathcal{Z}_{1:i-1}=[z_1, \cdots, z_{i-1}]$, the set of the next steps can be obtained by repeated sampling with the greedy method as $\{z_{ij}|z_{ij}\sim\pi_{\theta}(\mathcal{I}=\mathbb{I}(\mathcal{Q}, z_1, \cdots, z_{i-1}))\}$. 
Tree searching is an iterative generation process via repeatedly concatenating each sampled next step with a prefix sequence to form a new sequence before the next repeated sampling.
Thus, the tree generated from $z_{i-1}$ can be formed as $\mathcal{T}_{z_{i-1}}$.
In this tree, $z_{i-1}$ represents the root node in the first layer, and each node in the second layer is the child node of $z_{i-1}$ denoted as $\{z_{ij}\}_{j=1}^{N}$, which is also the root node in the corresponding tree $\mathcal{T}_{z_{ij}}$, $N$ is the number of repeated sampling at each layer.
% For the leaf node, which is the last reasoning step with the final answer
The searching process stops when the final answer is generated, and the last reasoning step can be viewed as the leaf node.
Similar to MCTS, each node has a corresponding value score denoted as $v(\cdot)$ that represents the potential that reaches the correct answer.
% In this paper, we follow the method in Math-Shepherd~\cite{Wang2024Math} to use the hard estimation to assign the value score, the value of the node $z_{ij}$ is represented as $v(z_{ij})\in\{0, 1\}$.

\begin{figure*}[t]
  \includegraphics[width=\linewidth]{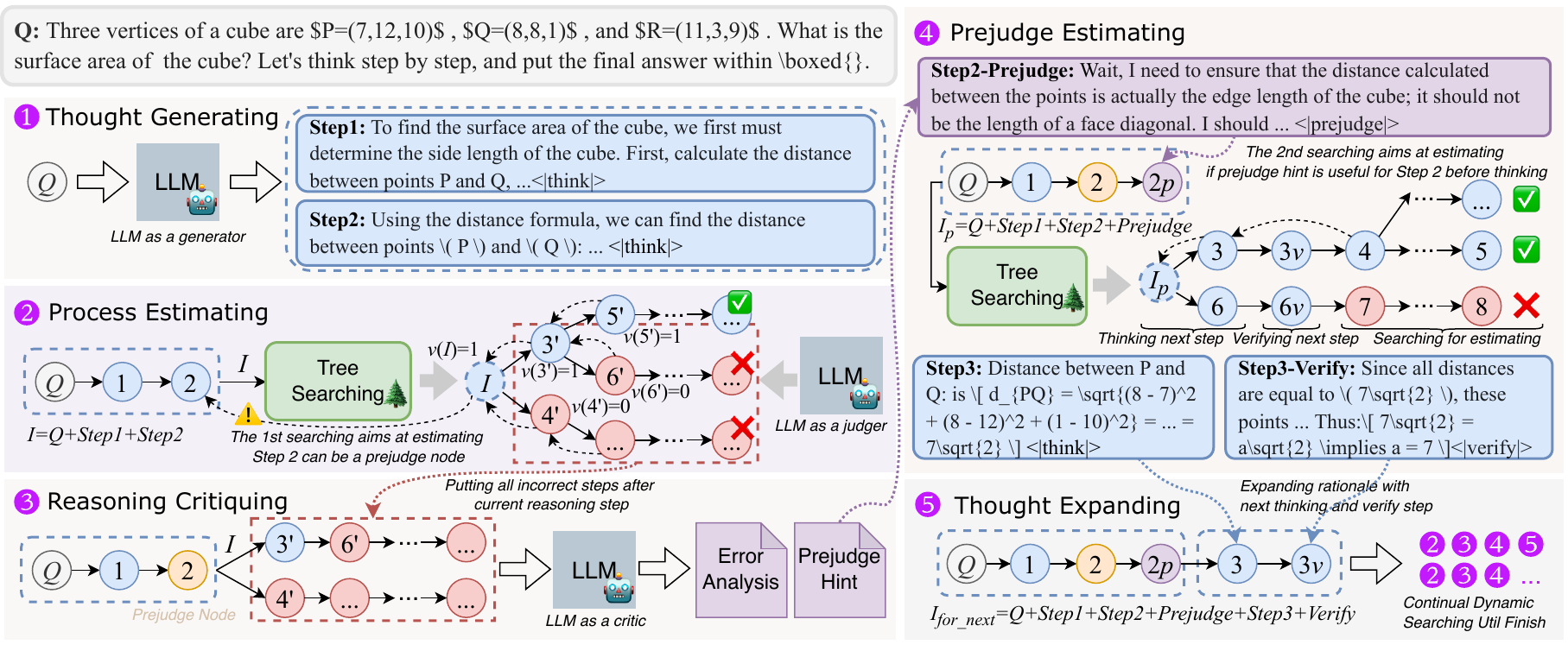} \hfill
  \caption {The automated reasoning framework for synthesizing process prejudge with the dynamic tree-searching.}
  \label{fig:framework}
\end{figure*}

\section{Methodology}

In this section, we introduce a new reasoning method named~\emph{Prejudge Before Think} (PBT), which aims to elicit the LLM pauses to deeply consider what errors will occur and how to bypass them instead of excessive trial and error.
Hence, we pose three questions to illustrate how our approach works:
\begin{itemize}
    \item \textbf{R1}: Where should the LLM pause to make a prejudgement when reasoning? 
    \item \textbf{R2}: How to obtain the trajectory with PBT automatically without any external annotation?
    \item \textbf{R3}: How to boost the LLM reasoning capability with post-training techniques?
\end{itemize}

\subsection{Prejudge Node in Rationale}

To answer the first question, we observe that humans can use historical experience to help them make prejudgements before facing potential risks~\cite{kahneman2011thinking}. Likewise, the time before errors occur is more suitable for stimulating the ability of LLMs to prejudge.
Hence, we define this position as~\emph{prejudge node}.
% , which is a specific step in the rationale.

Formally, given a question $\mathcal{Q}$ and the corresponding prefix reasoning steps $\mathcal{Z}_{1:i}$. To detect whether the LLM needs to make prejudgement after the step $z_{i}$, we can perform tree searching from this step to construct a tree denoted as $\mathcal{T}_{z_{i}}$, the value of each tree node can be obtained by hard estimation~\cite{Wang2024Math}.
Specifically, we first use LLM-as-a-judger to check whether the final answer is correct, and the value of each leaf node can be set as 1 (or 0) if the result is correct (or incorrect).
Then, the value of each internal node $z_{k}$ can be backtracked from the leaf node as:
\begin{equation}
v(z_{k}) = 
\begin{cases} 
% 1, & \text{if } k = T~\text{and}~z_{k}~\text{is correct} \\
% 0, & \text{if } k = T~\text{and}~z_{k}~\text{is incorrect} \\
\text{Judger}(z_k), & \text{if } k = T \\
max(\{v(z_{k+1j})\}_{j=1}^{N})), & \text{if } k < T 
\end{cases}
\label{eqn:hard_estimation}
\end{equation}
where $\text{Judger}(\cdot)\in\{0, 1\}$ is the function of LLM-as-a-judger, $z_{k+1j}$ is the child node of $z_{k}$.

Based on this value score, we can define a~\emph{prejudge value} that represents whether LLM should make prejudgement at the step $z_i$, which can be calculated as:
\begin{equation}
v_{p}(z_{i}) = v(z_{i})\times\mathbf{1}(min(\{v(z_{i+1j})\}_{j=1}^{N})=0),
\label{eqn:prejudge_score}
\end{equation}
where $v_p(\cdot)$ is the prejudge value and the step $z_i$ is a prejudge node when $v_p(z_i)=1$, $\mathbf{1}(\cdot)$ is the indicator function, $z_{i+1j}$ is the child node of $z_i$.
Through this definition, at least one step follows the prejudge node and has no path toward the correct answer. 
This indicates that the LLM may make errors in the future so that needs to be prejudged.

\subsection{Dynamic Tree Searching}
% 之前有篇工作发现先贪心生成一步，再进行搜索更好，所以我们每个第一步都只生成一步/两步
We thus introduce how to synthesize the reasoning data with prejudge.
We present a dynamic tree searching strategy in the reasoning framework, which enables only one LLM to find the prejudge position, generate critical information, and perform deep thought.
The whole framework is shown in Figure~\ref{fig:framework}, consisting of five stages: thought generating, process estimating, reasoning critiquing, prejudge estimating and thought expanding.

\noindent\paragraph{Thought Generating}
Given a question, we first to let the LLM generate a few steps without bootstrapping.
Inspired by~\cite{Wang2024Chain}, the start sequence has a greater impact on the performance of subsequent reasoning, especially for smaller models, so a lower temperature coefficient is selected to ensure the accuracy of reasoning in the first few steps.
By default, we urge the model to generate two steps as the prefix sequence. Each thinking step will be ended with a special tag as ``<|think|>''.

\noindent\paragraph{Process Estimating}
Once the prefix sequence is generated, we improve the temperature coefficient and perform the first tree-searching process.
This tree-searching aims to estimate whether the current step is a prejudge node, we can obtain the corresponding prejudge value $v_p$.
Specifically, if the prejudge value is $0$, it means that there is no need to make prejudgement here, we can randomly select one child node and concatenate it with the prefix sequence for the next iteration.
% Otherwise, if the prejudge value is $1$, the following process will proceed to generate critical information.
Otherwise, we will continue the following process to obtain the prejudge hint before the next thinking step.

\noindent\paragraph{Reasoning Critiquing}
This stage aims to generate error analysis based on the incorrect reasoning path derived from the tree search. Specifically, we concatenate the question and prefix sequence with all incorrect reasoning paths to form a prompt and use the LLM as a critic~\cite{Shinn2023Reflexion, Gou2024CRITIC} to generate the corresponding error analysis. Generally, error analysis summarizes and induces existing reasoning results, making all reasoning steps visible. In contrast, LLM prejudging guesses possible errors without examining subsequent reasoning steps. To achieve this goal, we design an instruction prompt for the LLM to generate a prejudging hint based on the error analysis, and this information will be used to prompt the LLM to avoid errors. The prejudging hint can be tagged with a ``<|prejudge|>'' token at the end of the text. The specific prompt and generated examples are shown in Appendix~\ref{app:prompt_for_critic}.

\noindent\paragraph{Prejudge Estimating}
After the prejudge hint is generated, we aim to evaluate its usefulness for the upcoming reasoning steps. Specifically, we perform tree-searching for the second time. Unlike the tree-searching conducted during the process estimation stage, the requirements of the generated response consist of three components: 
i) the next step thought (tagging with ``<|think|>''): we expect the LLM to reconsider the next reasoning step in light of the prejudge hint; 
ii) the verification of the next step thought (tagging with ``<|verify|>''): we introduce an explicit verification step for the LLM to check if the new thought aligns with the prejudge hint;
and iii) the remaining steps with bootstrapping (tagging with ``<|think|>''): we employ tree-searching to sample all reasoning steps to determine if the prejudge hint enables the LLM to arrive at the correct answer, utilizing the rejected sampling method to select the appropriate prejudge hint. We have crafted an instruction prompt to encourage the LLM to generate the aforementioned information, with the prompt and generated examples displayed in Appendix~\ref{app:prompt_for_prejudge_estimating}.

\noindent\paragraph{Thought Expanding}
Finally, we expand the reasoning chain with a prejudge hint, the selected next step consideration, and the corresponding verification. Then, we continue the next iteration of dynamic searching until we reach the final answer.

\subsection{Two-phase Post-training}

Dynamic tree searching involves significant test-time costs because it necessitates constructing multiple trees for each query. 
To equip the more efficient data synthesis pipeline with prejudge reasoning,
% To efficiently obtain large-scale synthesis data, 
we introduce a two-phase post-training strategy that allows for simultaneous data synthesis and model training.

\noindent\paragraph{First Phase: Cold Start via Dynamic Searching}
In the first phase, the aim is to construct a small amount of data with prejudge reasoning.
To collect this data, we choose a small instruct-based LLM to finish the dynamic tree-searching and obtain all correct rationales by rejected sampling.
Specifically, we select multiple training sources from GSM8K~\cite{Cobbe2021Training}, MATH~\cite{Hendrycks2021Measuring},  SVAMP~\cite{patel2021nlp}, AQuA~\cite{Ling2017Program}, Numina Math~\cite{numina_math_datasets}, American Invitational Mathematics Examination (AIME 1983$\sim$2023)~\footnote{\url{https://artofproblemsolving.com/wiki/index.php/AIME_Problems_and_Solutions}.}, PRM800K~\cite{Lightman2024Let}, and MetaMath QA~\cite{Yu2024MetaMath} and thus filter out about 21k complex queries for dynamic tree-searching, which can derive long CoT responses by zero-shot prompting~\cite{Kojima2022Large}.
Finally, we gather approximately 39k rationales and utilize these training samples to train a base LLM through SFT as the cold start model.

\noindent\paragraph{Second Phase: Distillation for Data Scaling}
The second phase focuses on self-evolution, aiming to leverage the SFT mode from the first phase to perform distillation. A large-scale, prejudge-style rationale will be constructed using a simple zero-shot prompting~\cite{Kojima2022Large}. 
To enhance the quality of the rationale, we employ the self-consistency strategy to recall the most reliable rationale, which can be utilized as the prejudge estimation stage in dynamic tree-searching. 
We thus select the remaining queries from the training sources, obtaining approximately 195k rationales. 
During the training periods, we combine the curated data generated from both phases and retrain SFT on the base LLM.
For the RL, we perform Group Relative Policy Optimization (GRPO) ~\cite{Shao2024DeepSeekMath} and Directly Preference Optimization (DPO)~\cite{Rafailov2023Direct} algorithms to investigate how performance improvement.

\begin{table*}[t]
\centering
\begin{small}
% \resizebox{\linewidth}{!}{
\begin{tabular}{l|cccccccc|c}
\toprule
\multirow{2}*{\textbf{Methods}} & \multirow{2}*{\textbf{GSM8K}} &\textbf{MATH} & \multirow{2}*{\textbf{AQuA}} & \multirow{2}*{\textbf{SVAMP}} & \textbf{Theorem}  & \textbf{AIME} & \textbf{GAOKAO} & \textbf{GPQA} &  \multirow{2}*{\textbf{Avg.}} \\
& & \bf 500 & & & \bf QA & \bf 2024 & \bf 2023 & \bf Diamond & \\
\midrule
OpenAI's o1 & - & 94.8 & - & - & - & 74.4 & - & 77.3 & -\\
GPT-4o & 94.2 & 76.8 & - & 93.9 & 49.7 & 9.3 & 88.1 & 50.6 & -\\
% QwQ-32B & - &  &  &  &  &  &  &  & - \\
DeepSeek R1 & - & 97.3 & - & - & - & 79.8 & - & 71.5 & - \\
\quad dist. Qwen-7B & - & 92.8 & - & - & - & 55.5 & - & 49.1 & - \\
\quad dist. Qwen-32B & - & 94.3 & - & - & - & 72.6 & - & 62.1 & - \\
\midrule
\multicolumn{10}{l}{\emph{Base Model: Qwen2.5-7B}} \\
\midrule
CoT$_{\text{\#1}}$ & 84.6 & 65.2 & 67.7 & 89.0 & 39.3 & 6.7 & 71.4 & 19.7 & 55.5 \\
Self-Refine$_{\text{\#1}}$ & 85.1 & 66.4 & 68.3 & 90.7 & 40.6 & 6.7 & 71.0 & 23.5 & 56.5 \\
\textbf{PBT}$_{\text{\#1}}$ & \bf 87.6 & \bf 68.0 & \bf 70.1 & 90.3 & \bf 41.4 & \bf 13.3 & \bf 73.5 & \bf 27.8 & \bf 59.0 \\
\quad w/o. Verify & 85.9 & 67.4 & 68.9 & 89.7 & 41.1 & 10.0 & 73.1 & 25.6 & 57.7 \\
\quad w. CoT & 85.6 & 67.2 & 68.5 & \bf 91.7 & 41.0 & 13.3 & 73.4 & 26.2 & 58.4 \\
\textbf{PBT}$_{\text{\#2}}$ & \bf 89.4 & \bf 72.6 & \bf 71.7 & \bf \bf 92.0 & \bf 43.2 & \bf 16.7 & \bf 75.5 & \bf 31.3 & \bf 61.6 \\

\midrule
\multicolumn{10}{l}{\emph{Base Model: Qwen2.5-32B}} \\
\midrule
CoT$_{\text{\#1}}$ & 91.7 & 77.6 & 75.2 & 89.3 & 48.7 & 6.7 & 77.6 & 38.4 & 63.2 \\
Self-Refine$_{\text{\#1}}$ & 92.6 & 77.3 & 76.8 & 90.4 & 49.3 & 10.0 & 79.4 & 39.4 & 64.4 \\
\textbf{PBT}$_{\text{\#1}}$ & 92.4 & \bf 78.2 & \bf 80.3 & 91.0 & 50.6 & 13.3 & 78.6 & \bf 40.2 & 65.6 \\
\quad w/o. Verify & 92.1 & 77.8 & 79.1 & 90.7 & 50.0 & 13.3 & 78.3 & 39.5 & 65.1 \\
\quad w. CoT & \bf 93.9 & 77.8 & 79.2 & \bf 92.3 & \bf 51.8 & \bf 16.7 & \bf 83.7 & 38.9 & \bf 66.8 \\
\textbf{PBT}$_{\text{\#2}}$ & \bf 93.1 & \bf 81.0 & \bf 81.7 & \bf 93.7 & \bf 52.0 & \bf 20.0 & \bf 85.6 & \bf 47.7 & \bf 69.4 \\

\bottomrule
\end{tabular}
% }
\end{small}
\caption{Main results (\%) over multiple complex reasoning tasks. \textbf{PBT} (prejudge before think) is our method, and the subscript $_{\text{\#1}}$ and $_{\text{\#2}}$ denotes the first phase and second phase, respectively.}
\label{tab:main_results}
\end{table*}

\section{Experiments}
\label{sec:experiments}
% In this section, we perform extensive experiments to evaluate our proposed method.

\subsection{Implementation Settings}
In the first phase, we choose Qwen2.5-14B-Instruct~\cite{team2024qwen2} as the small instruct-based LLM to perform dynamic tree-searching. By default, we use the ``$\backslash$n$\backslash$n'' as the step terminator, and the maximum sampling step length is 14 for each query. Complete implementation details of our dynamic tree-search algorithm are provided in Appendix~\ref{app:dynamic_tree_searching}.
For the cold start SFT, we choose Qwen2.5-7B/32B~\cite{team2024qwen2} as the base LLM.
In the second phase, we use the cold start 32B LLM to perform the distillation, and the number of repeated samples in Self-consistency is $N=32$. For the SFT and RL, we choose Qwen2.5-32B as the base model.
% More details of the SFT and RL are shown in Appendix~\ref{app:experiment_setup_details}.

\subsection{Benchmarks and Evaluations}
We select several competition-level reasoning benchmarks to demonstrate how LLM performance is improved with prejudge reasoning. These include GSM8K~\cite{Cobbe2021Training}, MATH-500~\cite{Lightman2024Let}, AQuA~\cite{Ling2017Program}, SVAMP~\cite{patel2021nlp}, TheoremQA~\cite{Chen2023TheoremQA}, AIME-2024~\cite{aime24}, GAOKAO-2023~\cite{Zhang2023Evaluating}, and GPQA-Diamond~\cite{Rein2023GPQA}.
We follow previous works~\cite{Wang2024Math} to use Qwen2.5-72B-Instruct~\cite{team2024qwen2} as a judger to evaluate whether the generated answer matches the ground truth, and the metric is accuracy value.

\subsection{Baselines}
We select the following baselines to compare with our method:
1) Chain-of-Thought (CoT) Training, which aims to distill multiple rationales using CoT prompts for the supervised fine-tuning (SFT) training data.
2) Self-Refine~\cite{Kumar2024Training}, which seeks to correct mistakes based on outcome- or process-based feedback. We obtain this rationale by prompting the small LLM with a well-designed zero-shot instruction.
3) PBT w/o. Verify, a variant version of our method that removes all verification components in dynamic tree searching; this means the LLM only makes prejudgments without any verification.
4) PBT w. CoT, which combines all data from prejudge and CoT.
We also select GPT-4o~\cite{Hurst2024GPT} and OpenAI's o1~\cite{Jaech2024OpenAI} as strong baselines to demonstrate state-of-the-art performance.

\subsection{Main Results}

Table~\ref{tab:main_results} presents the performance comparison with multiple baselines on competition-level complex reasoning tasks.
Through the results, we thus draw the following
conclusion:
1) Our PBT consistently outperforms CoT training and Self-Refine across all benchmarks using both 7B and 32B backbones. Specifically, in the first phase, PBT achieves an average accuracy of 59.0\% and 65.6\% for Qwen2.5-7B and Qwen2.5-32B, representing improvements of 3.5\% and 2.4\% over CoT and 2.5\% and 1.2\% over Self-Refine.
2) The ``verify'' components in PBT are crucial for ensuring the accuracy of prejudgment. We can see that all results decline when this component is removed, except for Qwen2.5-32B on AIME-2024.
3) Combining CoT data with PBT can further enhance the accuracy of the 32B model. We find that \textbf{PBT}$_{\text{\#1}}$ w. CoT outperforms \textbf{PBT}$_{\text{\#1}}$ by 1.2\%, indicating that larger models can benefit from diverse rationales, enabling them to utilize various styles of rationales to solve problems.
4) The two-phase strategy can bring obvious improvement. Compared with \textbf{PBT}$_{\text{\#1}}$, \textbf{PBT}$_{\text{\#2}}$ can further improve by 3\%, demonstrating the effectiveness of two-phase post-training strategy.
% In addition, it is worth noting that the model after two-phase training can get close to the level of GPT-4o by only using the data distilled from 14B and 32B models.

\begin{table}[t]
\centering
\begin{small}
\resizebox{\linewidth}{!}{
\begin{tabular}{l|ccc}
\toprule
\textbf{Methods} & \textbf{PBT}$_{\text{\#1}}$ & \bf w. DPO & \bf w. GRPO \\
\midrule
GSM8K & 87.6 & 90.5 & 91.7 \\
MATH500 & 68.0 & 70.4 & 71.2 \\
AQuA  & 70.1 & 74.0 & 74.8 \\
SVAMP  & 90.3 & 91.5 & 92.2 \\
Theorem QA  & 41.4 & 44.8 & 45.3 \\
AIME2024  & 13.3 & 16.7 & 26.7 \\
GAOKAO2023  & 73.5 & 78.4 & 79.3 \\
GPQA-Diamond  & 27.8 & 28.6 & 30.1 \\
\midrule
Avg.  & 59.0 & 61.9 & 63.9 \\
\bottomrule
\end{tabular}
}
\end{small}
\caption{The improvement (\%) of RL for Qwen2.5-7B.
}
\label{tab:rl_results}
\end{table}

\section{Analysis}
\subsection{Performance Improvement of RL}
To investigate performance improvements when applying RL with pre-judgment reasoning, we utilize two RL techniques to further enhance the SFT model in the first phase. Results shown in Table~\ref{tab:rl_results} demonstrate that the two RL methods substantially outperform the SFT. GRPO achieves the best performance, surpassing DPO by 2.0\%, indicating that online RL is more effective in boosting reasoning ability than the offline method.

\begin{figure}[t]
\centering
\begin{tabular}{cc}
\begin{minipage}[t]{0.49\linewidth}
    \includegraphics[width = 1\linewidth]{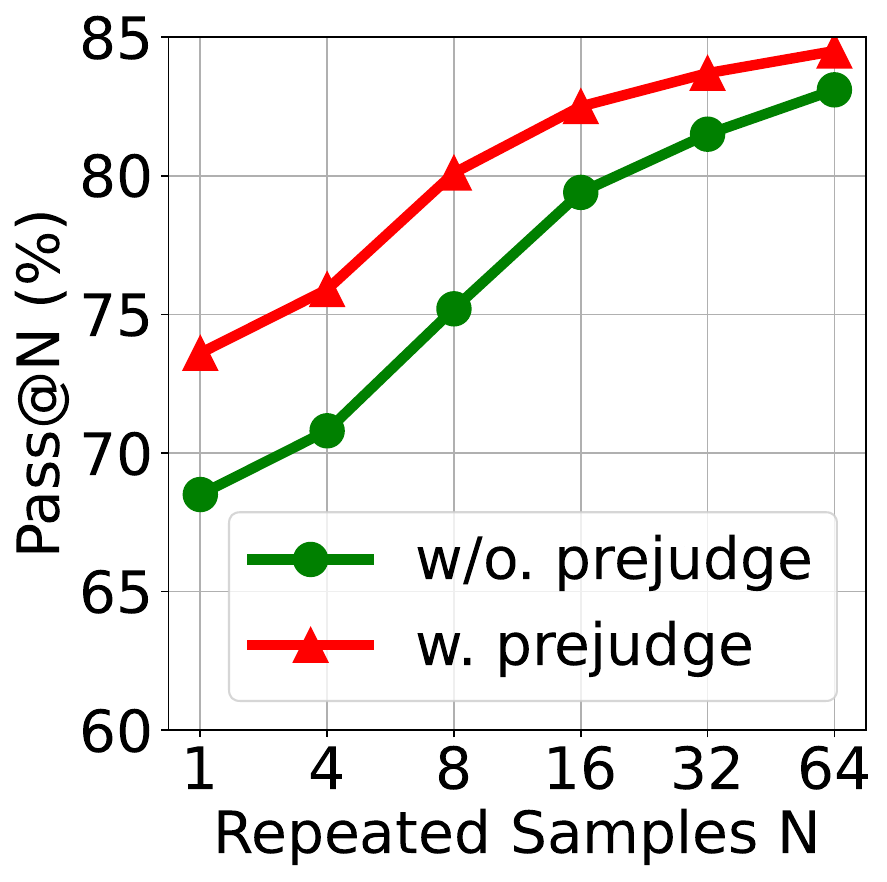}
\end{minipage}
\begin{minipage}[t]{0.49\linewidth}
    \includegraphics[width = 1\linewidth]{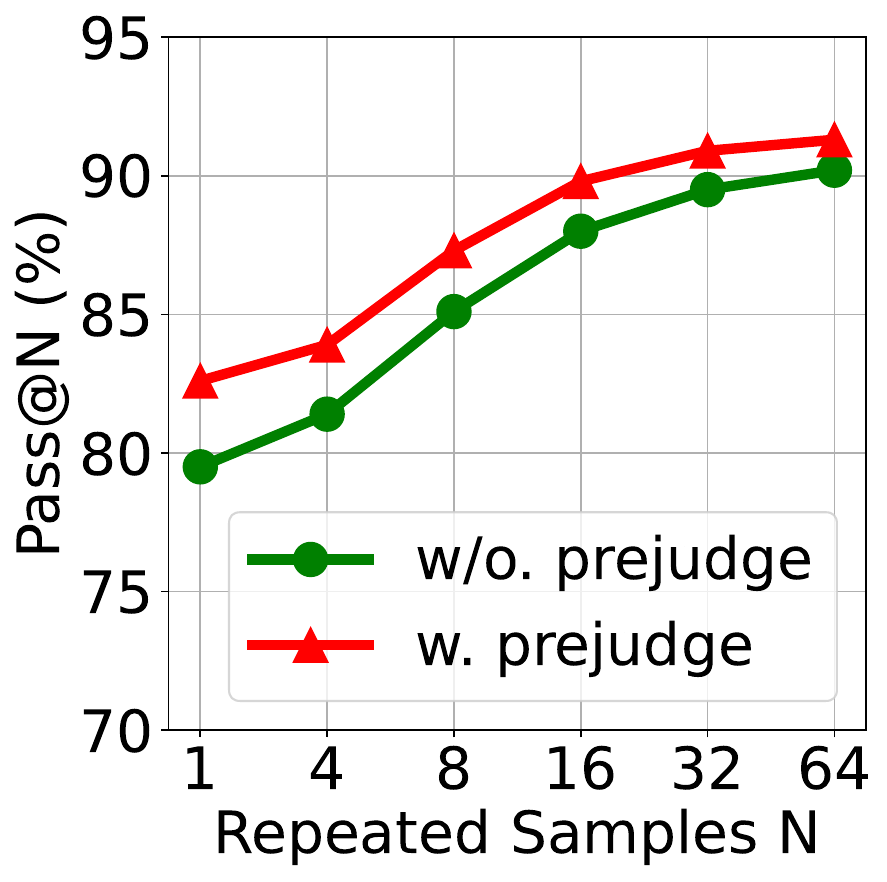}
\end{minipage}
\end{tabular}
\caption{Effect of prejudge in complex reasoning with Qwen2.5-7B-Instruct (Left) and Qwen2.5-32B-Instruct (Right).}
\label{fig:prejudge_effect}
\end{figure}

\subsection{Effect of Prejudge in Test-time}
To explore why prejudgment is effective for complex reasoning, we designed a thought completion task that demonstrates how the LLM reasons with prejudgment hints.
Specifically, we randomly select 2,000 queries from the second phase that do not appear in the first phase and keep only the first prejudge node with the prejudge hint and the prefix-generated steps.
For each query, we can obtain two incomplete responses: one that contains only the thinking step (i.e., remove the prejudge hint) and the other that contains the prejudge hint (i.e., maintain the prejudge hint).
We choose Qwen2.5-7B/32B-Instruct as the LLM.
To observe the performance, we let the LLM complete the reasoning steps with only the prompt ``Let's think step by step'' concatenated with the query and prefix-generated steps. We then draw a curve to see the performance when the test time scales up.
Figure~\ref{fig:prejudge_effect} shows that the Pass@N value is, on average, 3\% higher with prejudge hints than without, suggesting that the rationale provided by the prejudge hint can better assist the LLM in avoiding mistakes.

\begin{figure}[t]
\centering
\begin{tabular}{cc}
\begin{minipage}[t]{0.49\linewidth}
    \includegraphics[width = 1\linewidth]{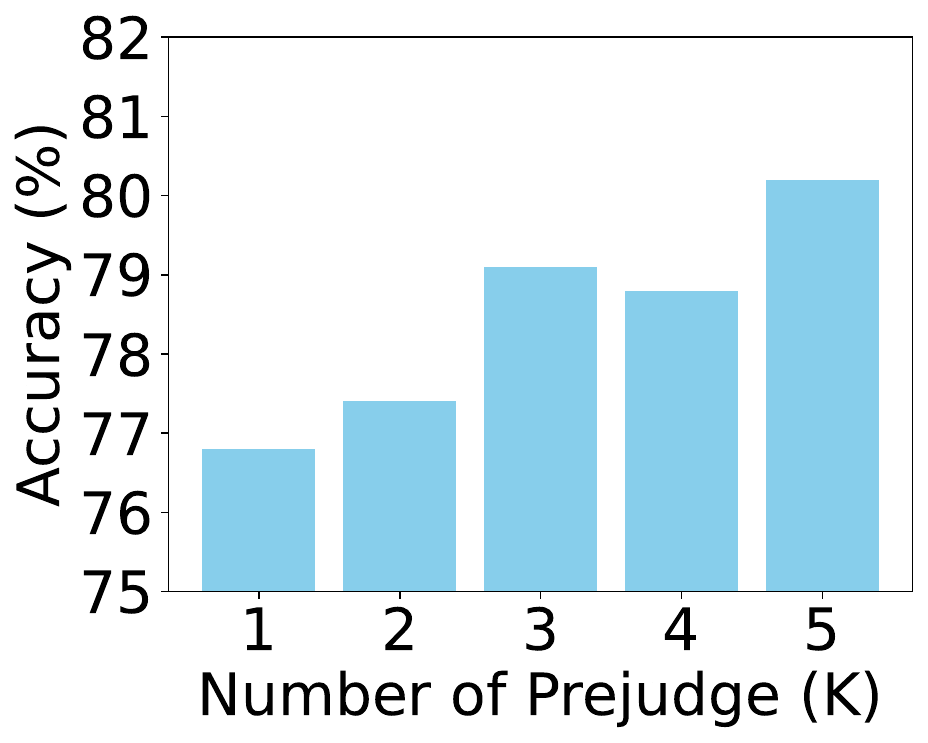}
\end{minipage}
\begin{minipage}[t]{0.49\linewidth}
    \includegraphics[width = 1\linewidth]{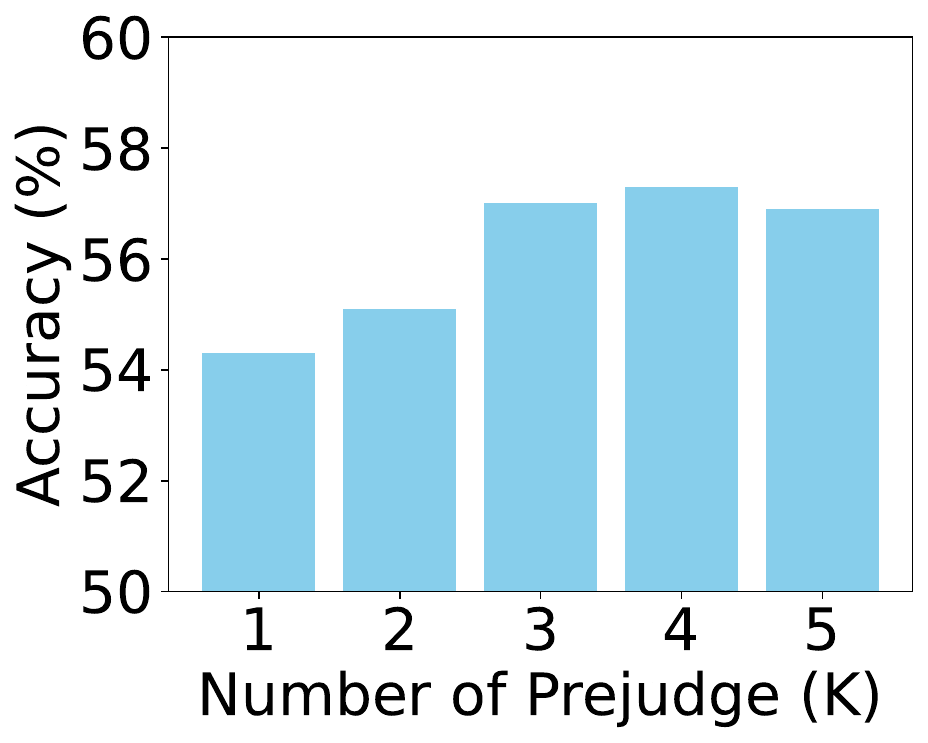}
\end{minipage}
\end{tabular}
\caption{Effect of the number of prejudge hints over GSM8K (Left) and MATH500 (Right).}
\label{fig:prejudge_number}
\end{figure}

\subsection{Effect of the Number of Prejudge}
In this section, we explore what's the effect of the number of prejudge in LLM reasoning.
We sample five different sets $\mathcal{D}_{k}$ from the synthesized data in the first phase for SFT training, where $k\in\{1, 2, 3, 4, 5\}$ denotes the number of the prejudge node in the corresponding rationale.
In other words, each rationale in $\mathcal{D}_{k}$ has only $k$ prejudge hint.
We perform data processing to ensure that the number of queries and rationales in each set is consistent.
To this end, each set has about 9k examples for SFT training.
The results displayed in Figure~\ref{fig:prejudge_number} suggest that adding the number of prejudge hints into the training data can significantly improve the reasoning ability of LLM.
We believe that the increase in the number of prejudges will indirectly increase the overall length of rationale, which can further improve the certainty of LLM's output when thinking about problems~\cite{Jaech2024OpenAI, guo2025deepseek}. 
In addition, the more prejudges there are, the more likely the model will make prejudges, which can better guide the model to make incorrect prejudges before thinking.

\begin{table}[t]
\centering
\begin{small}
\resizebox{\linewidth}{!}{
\begin{tabular}{l|ccc}
\toprule
\textbf{Methods} & \bf LIMO & \bf LIMO + PBT & \bf Gain \\
\midrule
GSM8K & 95.1 & 95.5 & +0.4 \\
MATH500 & 94.8 & 94.0 & -0.8 \\
AQuA  & 86.9 & 87.8 & +0.9 \\
SVAMP  & 91.3 & 92.3 & +1.0 \\
Theorem QA  & 54.6 & 58.0 & +4.6 \\
AIME2024  & 57.1 & 54.6 & -2.5 \\
GAOKAO2023  & 81.0 & 83.6 & +2.6 \\
GPQA-Diamond  & 66.7 & 63.0 & -3.7 \\
% Avg.  &  & - & - \\
\bottomrule
\end{tabular}
}
\end{small}
\caption{The performance (\%) with o1-like data.
}
\label{tab:o1_like_results}
\end{table}

\subsection{Compatibility with o1-like Reasoning}
We end this section by investigating the performance of mixing with o1-like training data. We select data that has been widely used recently from LIMO~\cite{ye2025limoreasoning}, which utilizes less data to achieve optimal performance on competition-level tasks. We gather all queries from AIME (1983$\sim$2023) to create prejudge reasoning data through dynamic tree searching and blend them with LIMO.
As shown in Table~\ref{tab:o1_like_results}, we can obtain the following suggestion:
Although prejudge before use is not entirely o1-like data, simply mixing them can still maintain very high performance, and some benchmarks can be further improved. This shows that prejudge can be better integrated into o1-like reasoning, which also provides a new mode as a reference for o1-like reasoning community.

\section{Related Works}

\noindent\paragraph{LLM Reasoning by Learning From Mistakes}
Developing the LLM with capabilities for correcting, reflecting, critiquing, and verifying has been one of the essential strategies for enhancing the LLM's reasoning ability. The essence of these methods is to learn from mistakes. Previous works aim to design zero-shot prompts or few-shot examples to encourage the LLM to utilize external feedback~\cite{Madaan2023Self, Welleck2023Generating, Xi2023Self, Wang2024Boosting}.
However, these methods heavily rely on external feedback and limit the model's ability to think spontaneously.
To remedy this dilemma, most recent works focus on post-training by injecting these abilities (i.e., correcting, reflecting, critiquing, and verifying) into model's parameters~\cite{Gao2024LLM, Wang2023Shepherd, Zhou2024Solving}.
Another line of research leverages self-training ways to develop these capabilities~\cite{Qu2024Recursive, Kumar2024Training, Zheng2024Critic, Xi2024Enhancing}.
Unlike them, we focus on the ability to prejudge, which helps the LLM take a moment to consider potential mistakes and think about how to avoid them before acting. Prejudging is also a way to learn from mistakes without trial and error.

\noindent\paragraph{Post-training in LLM Reasoning}
With the development of OpenAI's o1~\cite{Jaech2024OpenAI} and Deepseek R1~\cite{guo2025deepseek}, the post-training with test-time scaling has been powerful and versatile techniques in reasoning enhancement.
These studies typically increase inference
computation by extending the model’s thinking chains with tree search~\cite{Hao2023Reasoning, Zhang2024ReST, Zelikman2024Quiet, Nori2024From, Gao2024Interpretable}, process-based optimization~\cite{Uesato2022Solving, Wang2024Math, Lightman2024Let, Wang2024Q}, and self-play~\cite{Huang2023Large, Chen2024Self, Wang2024Self, Wu2024Self}.
To enhance the reasoning ability of small language models, some recent works perform distillation by DeepSeek R1 on Qwen2.5-7B and achieve satisfactory performance~\cite{Wen2025Light, ye2025limoreasoning}.
Unlike them, we use a small Instruct-like LLM and propose a dynamic tree-searching algorithm to synthesize rationale with prejudgment and verification and develop a two-phase post-training strategy to enhance the model's reasoning ability. We also investigate the performance gains achieved by scaling up the testing time, which indicates that prejudging before thinking can effectively elicit the model to avoid mistakes.

\section{Conclusion}

In this paper, we present a novel paradigm of "prejudge before thinking" inspired by System 2's slow thinking mode. We propose synthesizing training data using a dynamic tree-searching method with a small LLM and introduce a two-phase post-training strategy to enhance the model's reasoning ability with SFT and RL techniques. We conduct extensive experiments on multiple competition-level complex reasoning benchmarks, and the results demonstrate that the rationale embedded with the prejudged hints can guide the LLM to avoid making mistakes.

\section*{Limitations}

Our paper has some limitations, which we leave for
future work:

\noindent\paragraph{The computational cost of dynamic tree-searching}
The dynamic tree-searching algorithm incurs a high computational cost in the experiments. Each query takes about 5 minutes to generate four entire rationales. We attempted to use few-shot examples to prompt the strong LLM to generate the rationale with a preconceived notion, but we found that the preconceived notion was incorrect and the responses were not coherent.
In the future, we will focus on the time efficiency on the searching strategy.

\noindent\paragraph{The reasoning format}
We found that recent research from Deepseek R1 suggests that large-scale reinforcement learning based on a backbone can stimulate the model's self-reflection and slow thinking style. In contrast, our work focuses on data synthesis to construct System 2-like data. However, this leads us to a research topic concerning how to selectively activate specific reasoning modes through the RL stage. In other words, can we design a reward function or other strategies that allow LLMs to make prejudgments and combine them with some novel modes (e.g., aha moments) to enhance reasoning capabilities?

\section*{Social Impact and Ethics}

In terms of social impact,
the reasoning data we utilize are all from publicly available data sources. 
Infusing this information into the model's reasoning process will not introduce additional bias. 
% Moreover, it can to some extent prevent the model from providing irresponsible and harmful answers.
However, the open-source backbones we used may have some negative impacts, such as gender and social bias. 
Our work would unavoidably suffer from these issues.
We suggest that users should carefully address potential risks 
when the proposed method is deployed online.

\bibliography{anthology}

\begin{thebibliography}{62}
\providecommand{\natexlab}[1]{#1}

\bibitem[{Brown et~al.(2024)Brown, Juravsky, Ehrlich, Clark, Le, R{\'{e}}, and Mirhoseini}]{Brown2024Large}
Bradley C.~A. Brown, Jordan Juravsky, Ryan~Saul Ehrlich, Ronald Clark, Quoc~V. Le, Christopher R{\'{e}}, and Azalia Mirhoseini. 2024.
\newblock Large language monkeys: Scaling inference compute with repeated sampling.
\newblock \emph{CoRR}, abs/2407.21787.

\bibitem[{Brown et~al.(2020)Brown, Mann, Ryder, Subbiah, Kaplan, Dhariwal, Neelakantan, Shyam, Sastry, Askell, Agarwal, Herbert{-}Voss, Krueger, Henighan, Child, Ramesh, Ziegler, Wu, Winter, Hesse, Chen, Sigler, Litwin, Gray, Chess, Clark, Berner, McCandlish, Radford, Sutskever, and Amodei}]{Brown2020Language}
Tom~B. Brown, Benjamin Mann, Nick Ryder, Melanie Subbiah, Jared Kaplan, Prafulla Dhariwal, Arvind Neelakantan, Pranav Shyam, Girish Sastry, Amanda Askell, Sandhini Agarwal, Ariel Herbert{-}Voss, Gretchen Krueger, Tom Henighan, Rewon Child, Aditya Ramesh, Daniel~M. Ziegler, Jeffrey Wu, Clemens Winter, Christopher Hesse, Mark Chen, Eric Sigler, Mateusz Litwin, Scott Gray, Benjamin Chess, Jack Clark, Christopher Berner, Sam McCandlish, Alec Radford, Ilya Sutskever, and Dario Amodei. 2020.
\newblock Language models are few-shot learners.
\newblock In \emph{NeurIPS}.

\bibitem[{Chen and Li(2024)}]{Chen2024Toward}
Sijia Chen and Baochun Li. 2024.
\newblock Toward adaptive reasoning in large language models with thought rollback.
\newblock In \emph{{ICML}}. OpenReview.net.

\bibitem[{Chen et~al.(2023)Chen, Yin, Ku, Lu, Wan, Ma, Xu, Wang, and Xia}]{Chen2023TheoremQA}
Wenhu Chen, Ming Yin, Max Ku, Pan Lu, Yixin Wan, Xueguang Ma, Jianyu Xu, Xinyi Wang, and Tony Xia. 2023.
\newblock Theoremqa: {A} theorem-driven question answering dataset.
\newblock In \emph{{EMNLP}}, pages 7889--7901. Association for Computational Linguistics.

\bibitem[{Chen et~al.(2024)Chen, Deng, Yuan, Ji, and Gu}]{Chen2024Self}
Zixiang Chen, Yihe Deng, Huizhuo Yuan, Kaixuan Ji, and Quanquan Gu. 2024.
\newblock Self-play fine-tuning converts weak language models to strong language models.
\newblock In \emph{{ICML}}. OpenReview.net.

\bibitem[{Cobbe et~al.(2021)Cobbe, Kosaraju, Bavarian, Chen, Jun, Kaiser, Plappert, Tworek, Hilton, Nakano, Hesse, and Schulman}]{Cobbe2021Training}
Karl Cobbe, Vineet Kosaraju, Mohammad Bavarian, Mark Chen, Heewoo Jun, Lukasz Kaiser, Matthias Plappert, Jerry Tworek, Jacob Hilton, Reiichiro Nakano, Christopher Hesse, and John Schulman. 2021.
\newblock Training verifiers to solve math word problems.
\newblock \emph{CoRR}, abs/2110.14168.

\bibitem[{Gao et~al.(2024{\natexlab{a}})Gao, Cai, Xu, Wang, Zheng, Lin, Lu, Lin, Zhou, Xiao, Hu, Liu, and Chang}]{Gao2024LLM}
Bofei Gao, Zefan Cai, Runxin Xu, Peiyi Wang, Ce~Zheng, Runji Lin, Keming Lu, Junyang Lin, Chang Zhou, Wen Xiao, Junjie Hu, Tianyu Liu, and Baobao Chang. 2024{\natexlab{a}}.
\newblock {LLM} critics help catch bugs in mathematics: Towards a better mathematical verifier with natural language feedback.
\newblock \emph{CoRR}, abs/2406.14024.

\bibitem[{Gao et~al.(2024{\natexlab{b}})Gao, Niu, He, Xu, Liu, Liu, Hu, and Wen}]{Gao2024Interpretable}
Zitian Gao, Boye Niu, Xuzheng He, Haotian Xu, Hongzhang Liu, Aiwei Liu, Xuming Hu, and Lijie Wen. 2024{\natexlab{b}}.
\newblock Interpretable contrastive monte carlo tree search reasoning.
\newblock \emph{CoRR}, abs/2410.01707.

\bibitem[{Gou et~al.(2024)Gou, Shao, Gong, Shen, Yang, Duan, and Chen}]{Gou2024CRITIC}
Zhibin Gou, Zhihong Shao, Yeyun Gong, Yelong Shen, Yujiu Yang, Nan Duan, and Weizhu Chen. 2024.
\newblock {CRITIC:} large language models can self-correct with tool-interactive critiquing.
\newblock In \emph{{ICLR}}. OpenReview.net.

\bibitem[{Guo et~al.(2025)Guo, Yang, Zhang, Song, Zhang, Xu, Zhu, Ma, Wang, Bi et~al.}]{guo2025deepseek}
Daya Guo, Dejian Yang, Haowei Zhang, Junxiao Song, Ruoyu Zhang, Runxin Xu, Qihao Zhu, Shirong Ma, Peiyi Wang, Xiao Bi, et~al. 2025.
\newblock Deepseek-r1: Incentivizing reasoning capability in llms via reinforcement learning.
\newblock \emph{arXiv preprint arXiv:2501.12948}.

\bibitem[{Hao et~al.(2023)Hao, Gu, Ma, Hong, Wang, Wang, and Hu}]{Hao2023Reasoning}
Shibo Hao, Yi~Gu, Haodi Ma, Joshua~Jiahua Hong, Zhen Wang, Daisy~Zhe Wang, and Zhiting Hu. 2023.
\newblock Reasoning with language model is planning with world model.
\newblock In \emph{{EMNLP}}, pages 8154--8173. Association for Computational Linguistics.

\bibitem[{Hendrycks et~al.(2021)Hendrycks, Burns, Kadavath, Arora, Basart, Tang, Song, and Steinhardt}]{Hendrycks2021Measuring}
Dan Hendrycks, Collin Burns, Saurav Kadavath, Akul Arora, Steven Basart, Eric Tang, Dawn Song, and Jacob Steinhardt. 2021.
\newblock Measuring mathematical problem solving with the {MATH} dataset.
\newblock In \emph{NeurIPS}.

\bibitem[{Huang et~al.(2023)Huang, Gu, Hou, Wu, Wang, Yu, and Han}]{Huang2023Large}
Jiaxin Huang, Shixiang Gu, Le~Hou, Yuexin Wu, Xuezhi Wang, Hongkun Yu, and Jiawei Han. 2023.
\newblock Large language models can self-improve.
\newblock In \emph{{EMNLP}}, pages 1051--1068. Association for Computational Linguistics.

\bibitem[{Hurst et~al.(2024)Hurst, Lerer, Goucher, Perelman, Ramesh, Clark, Ostrow, Welihinda, Hayes, Radford, Madry, Baker{-}Whitcomb, Beutel, Borzunov, Carney, Chow, Kirillov, Nichol, Paino, Renzin, Passos, Kirillov, Christakis, Conneau, Kamali, Jabri, Moyer, Tam, Crookes, Tootoonchian, Kumar, Vallone, Karpathy, Braunstein, Cann, Codispoti, Galu, Kondrich, Tulloch, Mishchenko, Baek, Jiang, Pelisse, Woodford, Gosalia, Dhar, Pantuliano, Nayak, Oliver, Zoph, Ghorbani, Leimberger, Rossen, Sokolowsky, Wang, Zweig, Hoover, Samic, McGrew, Spero, Giertler, Cheng, Lightcap, Walkin, Quinn, Guarraci, Hsu, Kellogg, Eastman, Lugaresi, Wainwright, Bassin, Hudson, Chu, Nelson, Li, Shern, Conger, Barette, Voss, Ding, Lu, Zhang, Beaumont, Hallacy, Koch, Gibson, Kim, Choi, McLeavey, Hesse, Fischer, Winter, Czarnecki, Jarvis, Wei, Koumouzelis, and Sherburn}]{Hurst2024GPT}
Aaron Hurst, Adam Lerer, Adam~P. Goucher, Adam Perelman, Aditya Ramesh, Aidan Clark, AJ~Ostrow, Akila Welihinda, Alan Hayes, Alec Radford, Aleksander Madry, Alex Baker{-}Whitcomb, Alex Beutel, Alex Borzunov, Alex Carney, Alex Chow, Alex Kirillov, Alex Nichol, Alex Paino, Alex Renzin, Alex~Tachard Passos, Alexander Kirillov, Alexi Christakis, Alexis Conneau, Ali Kamali, Allan Jabri, Allison Moyer, Allison Tam, Amadou Crookes, Amin Tootoonchian, Ananya Kumar, Andrea Vallone, Andrej Karpathy, Andrew Braunstein, Andrew Cann, Andrew Codispoti, Andrew Galu, Andrew Kondrich, Andrew Tulloch, Andrey Mishchenko, Angela Baek, Angela Jiang, Antoine Pelisse, Antonia Woodford, Anuj Gosalia, Arka Dhar, Ashley Pantuliano, Avi Nayak, Avital Oliver, Barret Zoph, Behrooz Ghorbani, Ben Leimberger, Ben Rossen, Ben Sokolowsky, Ben Wang, Benjamin Zweig, Beth Hoover, Blake Samic, Bob McGrew, Bobby Spero, Bogo Giertler, Bowen Cheng, Brad Lightcap, Brandon Walkin, Brendan Quinn, Brian Guarraci, Brian Hsu, Bright Kellogg, Brydon
  Eastman, Camillo Lugaresi, Carroll~L. Wainwright, Cary Bassin, Cary Hudson, Casey Chu, Chad Nelson, Chak Li, Chan~Jun Shern, Channing Conger, Charlotte Barette, Chelsea Voss, Chen Ding, Cheng Lu, Chong Zhang, Chris Beaumont, Chris Hallacy, Chris Koch, Christian Gibson, Christina Kim, Christine Choi, Christine McLeavey, Christopher Hesse, Claudia Fischer, Clemens Winter, Coley Czarnecki, Colin Jarvis, Colin Wei, Constantin Koumouzelis, and Dane Sherburn. 2024.
\newblock Gpt-4o system card.
\newblock \emph{CoRR}, abs/2410.21276.

\bibitem[{Jaech et~al.(2024)Jaech, Kalai, Lerer, Richardson, El{-}Kishky, Low, Helyar, Madry, Beutel, Carney, Iftimie, Karpenko, Passos, Neitz, Prokofiev, Wei, Tam, Bennett, Kumar, Saraiva, Vallone, Duberstein, Kondrich, Mishchenko, Applebaum, Jiang, Nair, Zoph, Ghorbani, Rossen, Sokolowsky, Barak, McGrew, Minaiev, Hao, Baker, Houghton, McKinzie, Eastman, Lugaresi, Bassin, Hudson, Li, de~Bourcy, Voss, Shen, Zhang, Koch, Orsinger, Hesse, Fischer, Chan, Roberts, Kappler, Levy, Selsam, Dohan, Farhi, Mely, Robinson, Tsipras, Li, Oprica, Freeman, Zhang, Wong, Proehl, Cheung, Mitchell, Wallace, Ritter, Mays, Wang, Such, Raso, Leoni, Tsimpourlas, Song, von Lohmann, Sulit, Salmon, Parascandolo, Chabot, Zhao, Brockman, Leclerc, Salman, Bao, Sheng, Andrin, Bagherinezhad, Ren, Lightman, Chung, Kivlichan, O'Connell, Osband, Gilaberte, and Akkaya}]{Jaech2024OpenAI}
Aaron Jaech, Adam Kalai, Adam Lerer, Adam Richardson, Ahmed El{-}Kishky, Aiden Low, Alec Helyar, Aleksander Madry, Alex Beutel, Alex Carney, Alex Iftimie, Alex Karpenko, Alex~Tachard Passos, Alexander Neitz, Alexander Prokofiev, Alexander Wei, Allison Tam, Ally Bennett, Ananya Kumar, Andre Saraiva, Andrea Vallone, Andrew Duberstein, Andrew Kondrich, Andrey Mishchenko, Andy Applebaum, Angela Jiang, Ashvin Nair, Barret Zoph, Behrooz Ghorbani, Ben Rossen, Benjamin Sokolowsky, Boaz Barak, Bob McGrew, Borys Minaiev, Botao Hao, Bowen Baker, Brandon Houghton, Brandon McKinzie, Brydon Eastman, Camillo Lugaresi, Cary Bassin, Cary Hudson, Chak~Ming Li, Charles de~Bourcy, Chelsea Voss, Chen Shen, Chong Zhang, Chris Koch, Chris Orsinger, Christopher Hesse, Claudia Fischer, Clive Chan, Dan Roberts, Daniel Kappler, Daniel Levy, Daniel Selsam, David Dohan, David Farhi, David Mely, David Robinson, Dimitris Tsipras, Doug Li, Dragos Oprica, Eben Freeman, Eddie Zhang, Edmund Wong, Elizabeth Proehl, Enoch Cheung, Eric Mitchell,
  Eric Wallace, Erik Ritter, Evan Mays, Fan Wang, Felipe~Petroski Such, Filippo Raso, Florencia Leoni, Foivos Tsimpourlas, Francis Song, Fred von Lohmann, Freddie Sulit, Geoff Salmon, Giambattista Parascandolo, Gildas Chabot, Grace Zhao, Greg Brockman, Guillaume Leclerc, Hadi Salman, Haiming Bao, Hao Sheng, Hart Andrin, Hessam Bagherinezhad, Hongyu Ren, Hunter Lightman, Hyung~Won Chung, Ian Kivlichan, Ian O'Connell, Ian Osband, Ignasi~Clavera Gilaberte, and Ilge Akkaya. 2024.
\newblock Openai o1 system card.
\newblock \emph{CoRR}, abs/2412.16720.

\bibitem[{Kahneman(2011)}]{kahneman2011thinking}
Daniel Kahneman. 2011.
\newblock Thinking, fast and slow.
\newblock \emph{Farrar, Straus and Giroux}.

\bibitem[{Kocsis and Szepesv{\'{a}}ri(2006)}]{Kocsis2006Bandit}
Levente Kocsis and Csaba Szepesv{\'{a}}ri. 2006.
\newblock Bandit based monte-carlo planning.
\newblock In \emph{{ECML}}, volume 4212 of \emph{Lecture Notes in Computer Science}, pages 282--293. Springer.

\bibitem[{Kojima et~al.(2022)Kojima, Gu, Reid, Matsuo, and Iwasawa}]{Kojima2022Large}
Takeshi Kojima, Shixiang~Shane Gu, Machel Reid, Yutaka Matsuo, and Yusuke Iwasawa. 2022.
\newblock Large language models are zero-shot reasoners.
\newblock In \emph{NeurIPS}.

\bibitem[{Kumar et~al.(2024)Kumar, Zhuang, Agarwal, Su, Co{-}Reyes, Singh, Baumli, Iqbal, Bishop, Roelofs, Zhang, McKinney, Shrivastava, Paduraru, Tucker, Precup, Behbahani, and Faust}]{Kumar2024Training}
Aviral Kumar, Vincent Zhuang, Rishabh Agarwal, Yi~Su, John~D. Co{-}Reyes, Avi Singh, Kate Baumli, Shariq Iqbal, Colton Bishop, Rebecca Roelofs, Lei~M. Zhang, Kay McKinney, Disha Shrivastava, Cosmin Paduraru, George Tucker, Doina Precup, Feryal M.~P. Behbahani, and Aleksandra Faust. 2024.
\newblock Training language models to self-correct via reinforcement learning.
\newblock \emph{CoRR}, abs/2409.12917.

\bibitem[{LI et~al.(2024)LI, Beeching, Tunstall, Lipkin, Soletskyi, Huang, Rasul, Yu, Jiang, Shen, Qin, Dong, Zhou, Fleureau, Lample, and Polu}]{numina_math_datasets}
Jia LI, Edward Beeching, Lewis Tunstall, Ben Lipkin, Roman Soletskyi, Shengyi~Costa Huang, Kashif Rasul, Longhui Yu, Albert Jiang, Ziju Shen, Zihan Qin, Bin Dong, Li~Zhou, Yann Fleureau, Guillaume Lample, and Stanislas Polu. 2024.
\newblock Numinamath.

\bibitem[{Lightman et~al.(2024)Lightman, Kosaraju, Burda, Edwards, Baker, Lee, Leike, Schulman, Sutskever, and Cobbe}]{Lightman2024Let}
Hunter Lightman, Vineet Kosaraju, Yuri Burda, Harrison Edwards, Bowen Baker, Teddy Lee, Jan Leike, John Schulman, Ilya Sutskever, and Karl Cobbe. 2024.
\newblock Let's verify step by step.
\newblock In \emph{{ICLR}}. OpenReview.net.

\bibitem[{Ling et~al.(2017)Ling, Yogatama, Dyer, and Blunsom}]{Ling2017Program}
Wang Ling, Dani Yogatama, Chris Dyer, and Phil Blunsom. 2017.
\newblock Program induction by rationale generation: Learning to solve and explain algebraic word problems.
\newblock In \emph{{ACL}}, pages 158--167. Association for Computational Linguistics.

\bibitem[{Liu et~al.(2023)Liu, Yuan, Fu, Jiang, Hayashi, and Neubig}]{Liu2023Pre}
Pengfei Liu, Weizhe Yuan, Jinlan Fu, Zhengbao Jiang, Hiroaki Hayashi, and Graham Neubig. 2023.
\newblock Pre-train, prompt, and predict: {A} systematic survey of prompting methods in natural language processing.
\newblock \emph{{ACM} Comput. Surv.}, 55(9):195:1--195:35.

\bibitem[{MAA(2024)}]{aime24}
MAA. 2024.
\newblock American invitational mathematics examination - aime.
\newblock \emph{URL https://maa.org/math -competitions/american-invitational-mathematics-examination-aime}.

\bibitem[{Madaan et~al.(2023)Madaan, Tandon, Gupta, Hallinan, Gao, Wiegreffe, Alon, Dziri, Prabhumoye, Yang, Gupta, Majumder, Hermann, Welleck, Yazdanbakhsh, and Clark}]{Madaan2023Self}
Aman Madaan, Niket Tandon, Prakhar Gupta, Skyler Hallinan, Luyu Gao, Sarah Wiegreffe, Uri Alon, Nouha Dziri, Shrimai Prabhumoye, Yiming Yang, Shashank Gupta, Bodhisattwa~Prasad Majumder, Katherine Hermann, Sean Welleck, Amir Yazdanbakhsh, and Peter Clark. 2023.
\newblock Self-refine: Iterative refinement with self-feedback.
\newblock In \emph{NeurIPS}.

\bibitem[{Nori et~al.(2024)Nori, Usuyama, King, McKinney, Fernandes, Zhang, and Horvitz}]{Nori2024From}
Harsha Nori, Naoto Usuyama, Nicholas King, Scott~Mayer McKinney, Xavier Fernandes, Sheng Zhang, and Eric Horvitz. 2024.
\newblock From medprompt to o1: Exploration of run-time strategies for medical challenge problems and beyond.
\newblock \emph{CoRR}, abs/2411.03590.

\bibitem[{OpenAI(2023)}]{OpenAI2023GPT4}
OpenAI. 2023.
\newblock \href {https://doi.org/10.48550/arXiv.2303.08774} {{GPT-4} technical report}.
\newblock \emph{CoRR}, abs/2303.08774.

\bibitem[{Ouyang et~al.(2022)Ouyang, Wu, Jiang, Almeida, Wainwright, Mishkin, Zhang, Agarwal, Slama, Ray, Schulman, Hilton, Kelton, Miller, Simens, Askell, Welinder, Christiano, Leike, and Lowe}]{Ouyang2022Training}
Long Ouyang, Jeffrey Wu, Xu~Jiang, Diogo Almeida, Carroll~L. Wainwright, Pamela Mishkin, Chong Zhang, Sandhini Agarwal, Katarina Slama, Alex Ray, John Schulman, Jacob Hilton, Fraser Kelton, Luke Miller, Maddie Simens, Amanda Askell, Peter Welinder, Paul~F. Christiano, Jan Leike, and Ryan Lowe. 2022.
\newblock Training language models to follow instructions with human feedback.
\newblock In \emph{NeurIPS}.

\bibitem[{Park et~al.(2023)Park, O'Brien, Cai, Morris, Liang, and Bernstein}]{Park2023Generative}
Joon~Sung Park, Joseph~C. O'Brien, Carrie~Jun Cai, Meredith~Ringel Morris, Percy Liang, and Michael~S. Bernstein. 2023.
\newblock Generative agents: Interactive simulacra of human behavior.
\newblock In \emph{{UIST}}, pages 2:1--2:22. {ACM}.

\bibitem[{Patel et~al.(2021)Patel, Bhattamishra, and Goyal}]{patel2021nlp}
Arkil Patel, Satwik Bhattamishra, and Navin Goyal. 2021.
\newblock Are {NLP} models really able to solve simple math word problems?
\newblock In \emph{NAACL}, Online. Association for Computational Linguistics.

\bibitem[{Plaat et~al.(2024)Plaat, Wong, Verberne, Broekens, van Stein, and B{\"{a}}ck}]{Plaat2024Reasoning}
Aske Plaat, Annie Wong, Suzan Verberne, Joost Broekens, Niki van Stein, and Thomas B{\"{a}}ck. 2024.
\newblock Reasoning with large language models, a survey.
\newblock \emph{CoRR}, abs/2407.11511.

\bibitem[{Qi et~al.(2024)Qi, Ma, Xu, Zhang, Yang, and Yang}]{Qi2024Mutual}
Zhenting Qi, Mingyuan Ma, Jiahang Xu, Li~Lyna Zhang, Fan Yang, and Mao Yang. 2024.
\newblock Mutual reasoning makes smaller llms stronger problem-solvers.
\newblock \emph{CoRR}, abs/2408.06195.

\bibitem[{Qu et~al.(2024)Qu, Zhang, Garg, and Kumar}]{Qu2024Recursive}
Yuxiao Qu, Tianjun Zhang, Naman Garg, and Aviral Kumar. 2024.
\newblock Recursive introspection: Teaching language model agents how to self-improve.
\newblock In \emph{NeurIPS}.

\bibitem[{Rafailov et~al.(2023)Rafailov, Sharma, Mitchell, Ermon, Manning, and Finn}]{Rafailov2023Direct}
Rafael Rafailov, Archit Sharma, Eric Mitchell, Stefano Ermon, Christopher~D. Manning, and Chelsea Finn. 2023.
\newblock Direct preference optimization: Your language model is secretly a reward model.
\newblock \emph{CoRR}, abs/2305.18290.

\bibitem[{Rein et~al.(2023)Rein, Hou, Stickland, Petty, Pang, Dirani, Michael, and Bowman}]{Rein2023GPQA}
David Rein, Betty~Li Hou, Asa~Cooper Stickland, Jackson Petty, Richard~Yuanzhe Pang, Julien Dirani, Julian Michael, and Samuel~R. Bowman. 2023.
\newblock {GPQA:} {A} graduate-level google-proof q{\&}a benchmark.
\newblock \emph{CoRR}, abs/2311.12022.

\bibitem[{Shao et~al.(2024)Shao, Wang, Zhu, Xu, Song, Zhang, Li, Wu, and Guo}]{Shao2024DeepSeekMath}
Zhihong Shao, Peiyi Wang, Qihao Zhu, Runxin Xu, Junxiao Song, Mingchuan Zhang, Y.~K. Li, Y.~Wu, and Daya Guo. 2024.
\newblock Deepseekmath: Pushing the limits of mathematical reasoning in open language models.
\newblock \emph{CoRR}, abs/2402.03300.

\bibitem[{Shinn et~al.(2023)Shinn, Cassano, Gopinath, Narasimhan, and Yao}]{Shinn2023Reflexion}
Noah Shinn, Federico Cassano, Ashwin Gopinath, Karthik Narasimhan, and Shunyu Yao. 2023.
\newblock Reflexion: language agents with verbal reinforcement learning.
\newblock In \emph{NeurIPS}.

\bibitem[{Silver et~al.(2016)Silver, Huang, Maddison, Guez, Sifre, van~den Driessche, Schrittwieser, Antonoglou, Panneershelvam, Lanctot, Dieleman, Grewe, Nham, Kalchbrenner, Sutskever, Lillicrap, Leach, Kavukcuoglu, Graepel, and Hassabis}]{Silver2016Mastering}
David Silver, Aja Huang, Chris~J. Maddison, Arthur Guez, Laurent Sifre, George van~den Driessche, Julian Schrittwieser, Ioannis Antonoglou, Vedavyas Panneershelvam, Marc Lanctot, Sander Dieleman, Dominik Grewe, John Nham, Nal Kalchbrenner, Ilya Sutskever, Timothy~P. Lillicrap, Madeleine Leach, Koray Kavukcuoglu, Thore Graepel, and Demis Hassabis. 2016.
\newblock Mastering the game of go with deep neural networks and tree search.
\newblock \emph{Nat.}, 529(7587):484--489.

\bibitem[{Snell et~al.(2024)Snell, Lee, Xu, and Kumar}]{Snell2024Scaling}
Charlie Snell, Jaehoon Lee, Kelvin Xu, and Aviral Kumar. 2024.
\newblock Scaling {LLM} test-time compute optimally can be more effective than scaling model parameters.
\newblock \emph{CoRR}, abs/2408.03314.

\bibitem[{Team(2024)}]{team2024qwen2}
Qwen Team. 2024.
\newblock Qwen2. 5: A party of foundation models.
\newblock \emph{Qwen (Sept. 2024). url: https://qwenlm. github. io/blog/qwen2}, 5.

\bibitem[{Uesato et~al.(2022)Uesato, Kushman, Kumar, Song, Siegel, Wang, Creswell, Irving, and Higgins}]{Uesato2022Solving}
Jonathan Uesato, Nate Kushman, Ramana Kumar, H.~Francis Song, Noah~Y. Siegel, Lisa Wang, Antonia Creswell, Geoffrey Irving, and Irina Higgins. 2022.
\newblock Solving math word problems with process- and outcome-based feedback.
\newblock \emph{CoRR}, abs/2211.14275.

\bibitem[{Wang et~al.(2024{\natexlab{a}})Wang, Deng, Lv, Liang, He, Yan, and An}]{Wang2024Q}
Chaojie Wang, Yanchen Deng, Zhiyi Lv, Zeng Liang, Jujie He, Shuicheng Yan, and Bo~An. 2024{\natexlab{a}}.
\newblock Q*: Improving multi-step reasoning for llms with deliberative planning.
\newblock \emph{CoRR}, abs/2406.14283.

\bibitem[{Wang et~al.(2024{\natexlab{b}})Wang, Sun, Li, and Gao}]{Wang2024Boosting}
Jianing Wang, Qiushi Sun, Xiang Li, and Ming Gao. 2024{\natexlab{b}}.
\newblock Boosting language models reasoning with chain-of-knowledge prompting.
\newblock In \emph{ACL}, pages 4958--4981. Association for Computational Linguistics.

\bibitem[{Wang et~al.(2024{\natexlab{c}})Wang, Wu, Hou, Liu, Gao, and McAuley}]{Wang2024InstructGraph}
Jianing Wang, Junda Wu, Yupeng Hou, Yao Liu, Ming Gao, and Julian~J. McAuley. 2024{\natexlab{c}}.
\newblock Instructgraph: Boosting large language models via graph-centric instruction tuning and preference alignment.
\newblock In \emph{{ACL}}, pages 13492--13510. Association for Computational Linguistics.

\bibitem[{Wang et~al.(2024{\natexlab{d}})Wang, Zhou, Zhang, Bao, and Yan}]{Wang2024Self}
Jianing Wang, Yang Zhou, Xiaocheng Zhang, Mengjiao Bao, and Peng Yan. 2024{\natexlab{d}}.
\newblock Self-evolutionary large language models through uncertainty-enhanced preference optimization.
\newblock \emph{CoRR}, abs/2409.11212.

\bibitem[{Wang et~al.(2024{\natexlab{e}})Wang, Li, Shao, Xu, Dai, Li, Chen, Wu, and Sui}]{Wang2024Math}
Peiyi Wang, Lei Li, Zhihong Shao, Runxin Xu, Damai Dai, Yifei Li, Deli Chen, Yu~Wu, and Zhifang Sui. 2024{\natexlab{e}}.
\newblock Math-shepherd: Verify and reinforce llms step-by-step without human annotations.
\newblock In \emph{{ACL}}, pages 9426--9439. Association for Computational Linguistics.

\bibitem[{Wang et~al.(2023{\natexlab{a}})Wang, Yu, Tan, O'Brien, Pasunuru, Dwivedi{-}Yu, Golovneva, Zettlemoyer, Fazel{-}Zarandi, and Celikyilmaz}]{Wang2023Shepherd}
Tianlu Wang, Ping Yu, Xiaoqing~Ellen Tan, Sean O'Brien, Ramakanth Pasunuru, Jane Dwivedi{-}Yu, Olga Golovneva, Luke Zettlemoyer, Maryam Fazel{-}Zarandi, and Asli Celikyilmaz. 2023{\natexlab{a}}.
\newblock Shepherd: {A} critic for language model generation.
\newblock \emph{CoRR}, abs/2308.04592.

\bibitem[{Wang et~al.(2023{\natexlab{b}})Wang, Wei, Schuurmans, Le, Chi, Narang, Chowdhery, and Zhou}]{Wang2023Self}
Xuezhi Wang, Jason Wei, Dale Schuurmans, Quoc~V. Le, Ed~H. Chi, Sharan Narang, Aakanksha Chowdhery, and Denny Zhou. 2023{\natexlab{b}}.
\newblock Self-consistency improves chain of thought reasoning in language models.
\newblock In \emph{{ICLR}}. OpenReview.net.

\bibitem[{Wang and Zhou(2024)}]{Wang2024Chain}
Xuezhi Wang and Denny Zhou. 2024.
\newblock Chain-of-thought reasoning without prompting.
\newblock In \emph{NeurIPS}.

\bibitem[{Wei et~al.(2022)Wei, Wang, Schuurmans, Bosma, Ichter, Xia, Chi, Le, and Zhou}]{Wei2022Chain}
Jason Wei, Xuezhi Wang, Dale Schuurmans, Maarten Bosma, Brian Ichter, Fei Xia, Ed~H. Chi, Quoc~V. Le, and Denny Zhou. 2022.
\newblock Chain-of-thought prompting elicits reasoning in large language models.
\newblock In \emph{NeurIPS}.

\bibitem[{Welleck et~al.(2023)Welleck, Lu, West, Brahman, Shen, Khashabi, and Choi}]{Welleck2023Generating}
Sean Welleck, Ximing Lu, Peter West, Faeze Brahman, Tianxiao Shen, Daniel Khashabi, and Yejin Choi. 2023.
\newblock Generating sequences by learning to self-correct.
\newblock In \emph{{ICLR}}. OpenReview.net.

\bibitem[{Wen et~al.(2025)Wen, Cai, Xiao, He, An, Duan, Du, Liu, Tang, Lv, Zou, Deng, Jia, and Zhang}]{Wen2025Light}
Liang Wen, Yunke Cai, Fenrui Xiao, Xin He, Qi~An, Zhenyu Duan, Yimin Du, Junchen Liu, Lifu Tang, Xiaowei Lv, Haosheng Zou, Yongchao Deng, Shousheng Jia, and Xiangzheng Zhang. 2025.
\newblock \href {https://arxiv.org/abs/2503.10460} {Light-r1: Curriculum sft, dpo and rl for long cot from scratch and beyond}.
\newblock \emph{Preprint}, arXiv:2503.10460.

\bibitem[{Wu et~al.(2024)Wu, Sun, Yuan, Ji, Yang, and Gu}]{Wu2024Self}
Yue Wu, Zhiqing Sun, Huizhuo Yuan, Kaixuan Ji, Yiming Yang, and Quanquan Gu. 2024.
\newblock Self-play preference optimization for language model alignment.
\newblock \emph{CoRR}, abs/2405.00675.

\bibitem[{Xi et~al.(2023)Xi, Jin, Zhou, Zheng, Gao, Liu, Gui, Zhang, and Huang}]{Xi2023Self}
Zhiheng Xi, Senjie Jin, Yuhao Zhou, Rui Zheng, Songyang Gao, Jia Liu, Tao Gui, Qi~Zhang, and Xuanjing Huang. 2023.
\newblock Self-polish: Enhance reasoning in large language models via problem refinement.
\newblock In \emph{{EMNLP}}, pages 11383--11406. Association for Computational Linguistics.

\bibitem[{Xi et~al.(2024)Xi, Yang, Huang, Tang, Li, Ding, He, Hong, Dou, Zhan, Wang, Zheng, Ji, Shi, Zhai, Weng, Wang, Cai, Gui, Wu, Zhang, Qiu, Huang, and Jiang}]{Xi2024Enhancing}
Zhiheng Xi, Dingwen Yang, Jixuan Huang, Jiafu Tang, Guanyu Li, Yiwen Ding, Wei He, Boyang Hong, Shihan Dou, Wenyu Zhan, Xiao Wang, Rui Zheng, Tao Ji, Xiaowei Shi, Yitao Zhai, Rongxiang Weng, Jingang Wang, Xunliang Cai, Tao Gui, Zuxuan Wu, Qi~Zhang, Xipeng Qiu, Xuanjing Huang, and Yu{-}Gang Jiang. 2024.
\newblock Enhancing {LLM} reasoning via critique models with test-time and training-time supervision.
\newblock \emph{CoRR}, abs/2411.16579.

\bibitem[{Ye et~al.(2025)Ye, Huang, Xiao, Chern, Xia, and Liu}]{ye2025limoreasoning}
Yixin Ye, Zhen Huang, Yang Xiao, Ethan Chern, Shijie Xia, and Pengfei Liu. 2025.
\newblock Limo: Less is more for reasoning.
\newblock In \emph{arXiv}. arXiv.

\bibitem[{Yu et~al.(2024)Yu, Jiang, Shi, Yu, Liu, Zhang, Kwok, Li, Weller, and Liu}]{Yu2024MetaMath}
Longhui Yu, Weisen Jiang, Han Shi, Jincheng Yu, Zhengying Liu, Yu~Zhang, James~T. Kwok, Zhenguo Li, Adrian Weller, and Weiyang Liu. 2024.
\newblock Metamath: Bootstrap your own mathematical questions for large language models.
\newblock In \emph{{ICLR}}. OpenReview.net.

\bibitem[{Zelikman et~al.(2024)Zelikman, Harik, Shao, Jayasiri, Haber, and Goodman}]{Zelikman2024Quiet}
Eric Zelikman, Georges Harik, Yijia Shao, Varuna Jayasiri, Nick Haber, and Noah~D. Goodman. 2024.
\newblock Quiet-star: Language models can teach themselves to think before speaking.
\newblock \emph{CoRR}, abs/2403.09629.

\bibitem[{Zhang et~al.(2024)Zhang, Zhoubian, Hu, Yue, Dong, and Tang}]{Zhang2024ReST}
Dan Zhang, Sining Zhoubian, Ziniu Hu, Yisong Yue, Yuxiao Dong, and Jie Tang. 2024.
\newblock Rest-mcts*: {LLM} self-training via process reward guided tree search.
\newblock In \emph{NeurIPS}.

\bibitem[{Zhang et~al.(2023)Zhang, Li, Zong, Ying, He, and Qiu}]{Zhang2023Evaluating}
Xiaotian Zhang, Chunyang Li, Yi~Zong, Zhengyu Ying, Liang He, and Xipeng Qiu. 2023.
\newblock Evaluating the performance of large language models on {GAOKAO} benchmark.
\newblock \emph{CoRR}, abs/2305.12474.

\bibitem[{Zheng et~al.(2024)Zheng, Lou, Cao, Wen, Ji, Lin, Lu, Han, Zhang, and Sun}]{Zheng2024Critic}
Xin Zheng, Jie Lou, Boxi Cao, Xueru Wen, Yuqiu Ji, Hongyu Lin, Yaojie Lu, Xianpei Han, Debing Zhang, and Le~Sun. 2024.
\newblock Critic-cot: Boosting the reasoning abilities of large language model via chain-of-thoughts critic.
\newblock \emph{CoRR}, abs/2408.16326.

\bibitem[{Zhou et~al.(2024)Zhou, Wang, Lu, Shi, Luo, Qin, Lu, Jia, Song, Zhan, and Li}]{Zhou2024Solving}
Aojun Zhou, Ke~Wang, Zimu Lu, Weikang Shi, Sichun Luo, Zipeng Qin, Shaoqing Lu, Anya Jia, Linqi Song, Mingjie Zhan, and Hongsheng Li. 2024.
\newblock Solving challenging math word problems using {GPT-4} code interpreter with code-based self-verification.
\newblock In \emph{{ICLR}}. OpenReview.net.

\end{thebibliography}
% \bibliography{custom}

\newpage

\appendix

\begin{table*}[t]
\centering
% \resizebox{\linewidth}{!}{
\begin{small}
\begin{tabular}{cc cc ccc c}
\toprule
\multirow{2}*{\textbf{Task}} & \multirow{2}*{\textbf{Domain}} & \multirow{2}*{\textbf{Source}} & \multirow{2}*{\textbf{Sampling}} & \multicolumn{2}{c}{\textbf{The 1st Phase}} & \multicolumn{2}{c}{\textbf{The 2nd Phase}}  \\
&  &  &  &\bf \textbf{\#Search} &\bf \textbf{\#Train} & \bf \textbf{\#Search} & \bf \textbf{\#Train} \\
\hline
GSM8K & MATH & \cite{Cobbe2021Training} & 7,473 & 7,473 & 7,473 & 0 & 0 \\
MATH & MATH & \cite{Hendrycks2021Measuring} & 3,994 & 3,994 & 2,000 & 1,994 & 1,200  \\
SVAMP & MATH & \cite{patel2021nlp} & 700 & 700 & 700 & 0 & 0  \\
AQuA & MATH & \cite{Ling2017Program} & 97467 & 7,000 & 5,350 & 1,650 & 210  \\
Numina Math & MATH & \cite{numina_math_datasets} & 34,473 & 12,000 & 4,800 & 22,473 & 15,800  \\
AIME (1983$\sim$2023) & MATH & \cite{numina_math_datasets} & 919 & 919 & 678 & 241 & 89  \\
PRM800K & MATH & \cite{Lightman2024Let} & 12,000 & 12,000 & 7,800 & 4,200 &  1,390 \\
MetaMath QA & MATH & \cite{Yu2024MetaMath} & 150,000 & 0 & 0 & 150,000 & 63,077 \\
\bottomrule
\end{tabular}
\end{small}
% }
\caption{The data statistics of each task. The data for \#Train is smaller than \#Search because rejected sampling.}
\label{tab:corpus}
\end{table*}

\section{Data Sources}
\label{app:data_sources}

For data collection, we select multiple training sources from GSM8K~\cite{Cobbe2021Training}, MATH~\cite{Hendrycks2021Measuring},  SVAMP~\cite{patel2021nlp}, AQuA~\cite{Ling2017Program}, Numina Math~\cite{numina_math_datasets}, American Invitational Mathematics Examination (AIME 1983$\sim$2023)~\footnote{\url{https://artofproblemsolving.com/wiki/index.php/AIME_Problems_and_Solutions}.}, PRM800K~\cite{Lightman2024Let}, and MetaMath QA~\cite{Yu2024MetaMath}.
The details of each source is shown in Table~\ref{tab:corpus}.

\section{Experimental Setup Details}
\label{app:experiment_setup_details}

\subsection{Details of Dynamic Tree-Searching}
\label{app:dynamic_tree_searching}
We develop a dynamic tree-searching to release process estimation and prejudge estimation.
We first numbered and named the internal nodes of each tree. In order to facilitate tracing each node, we adopted a continuous coding strategy.
For example, the node ``2-4-1-3'' is located at the fourth layer in the tree, and it is one of the child nodes of ``2-4-1''. the reasoning step at the node ``2-4-1-3'' can be viewed as the third repeated sample generated from ``2-4-1''.

Since tree search is an algorithm with exponentially increasing complexity, we agree that the number of repeated samplings at each layer is different and, finally, ensure that the number of paths (from root to all leaf nodes) does not exceed 1024.

% \subsection{Details of SFT}
% \label{app:sft}

% \subsection{Details of DPO}
% \label{app:dpo}

% \subsection{Details of GRPO}
% \label{app:grpo}

\section{Prompt Engineering}
\label{app:prompt_enginnering}

\subsection{Prompt Format}
\label{app:prompt_format}
In this paper, we choose Qwen2.5-7B and Qwen2.5-32B as the backbone for post-training, so we design a new prompt format for the subsequence training. The format is: 
\begin{verbatim}
<|im_start|>user
{Question}
Let's think step by step, and put
the final answer within \boxed{}.
<|im_end|>
<|im_start|>assistant
\end{verbatim}
% ``<|im\_start|>user$\backslash$n\{Question\}$\backslash$nLet's think step by step, and put the final answer within \\boxed\{\}.$\backslash$n<|im\_end|>$\backslash$n<|im\_start|>assistant$\backslash$n'',
where ``\{Question\}'' is the placeholder for complex query, ``<|im\_start|>'', ``<|im\_end|>'' are the special tokens in vocabulary set of Qwen2.5 model.

\subsection{Prompt for LLM-as-a-Judger}
\label{app:prompt_for_judger}

\begin{figure}[t]
  \includegraphics[width=\linewidth]{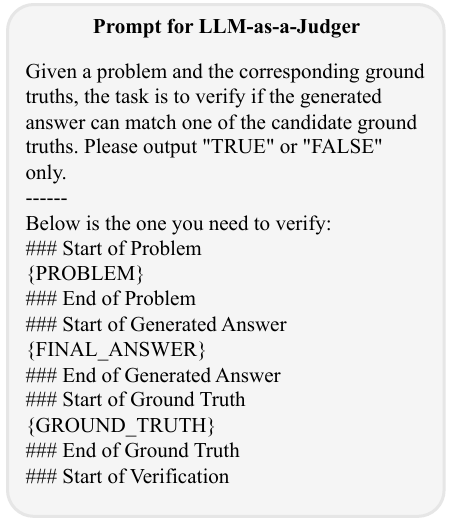} \hfill
  \caption {The prompt for LLM-as-a-Judger.}
  \label{fig:prompt_judger}
\end{figure}

The prompt for making the LLM-as-a-Judger is shown in Figure~\ref{fig:prompt_judger}.
The prompt will be used in dynamic tree searching to detect whether each reasoning path is correct (i.e., matching the final answer in the box with the ground truth). 

\begin{figure}[t]
  \includegraphics[width=\linewidth]{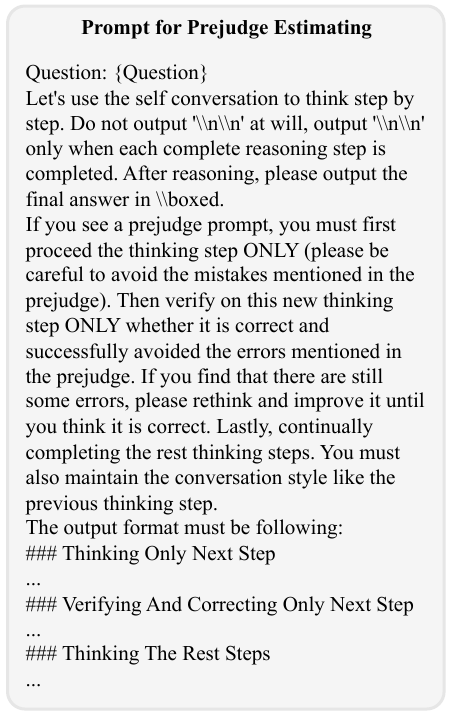} \hfill
  \caption {The prompt for prejudge estimating.}
  \label{fig:prompt_prejudge_estimating}
\end{figure}

\subsection{Prompt for LLM-as-a-Critic}
\label{app:prompt_for_critic}
\begin{figure*}[t]
  \includegraphics[width=\linewidth]{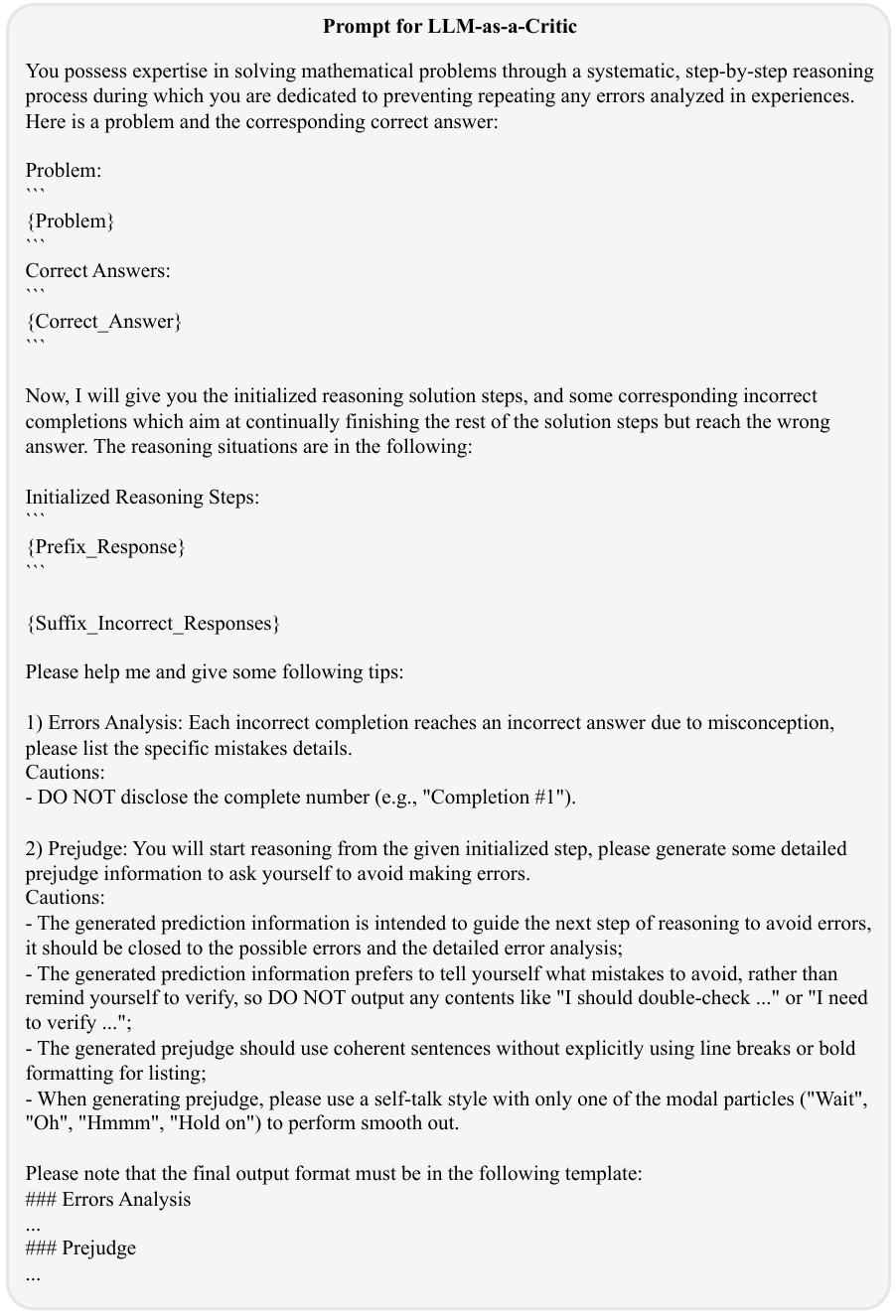} \hfill
  \caption {The prompt for LLM-as-a-Critic.}
  \label{fig:prompt_critic}
\end{figure*}

The prompt for generating analysis and prejudge hint is shown in Figure~\ref{fig:prompt_critic}. This prompt will be utilized in dynamic tree searching to produce error analysis for all incorrect rationales and construct a prejudge hint for the prejudge node.

\subsection{Prompt for Prejudge Estimating}
\label{app:prompt_for_prejudge_estimating}

The prompt for prejudge estimating is shown in Figure~\ref{fig:prompt_prejudge_estimating}. When the prejudge hint is generated, we can use this prompt to elicit the LLM to generate the next thinking step, verify that step, and proceed with the remaining steps to reach the final answer.

\section{Case Study}
\begin{figure*}[t]
  \includegraphics[width=\linewidth]{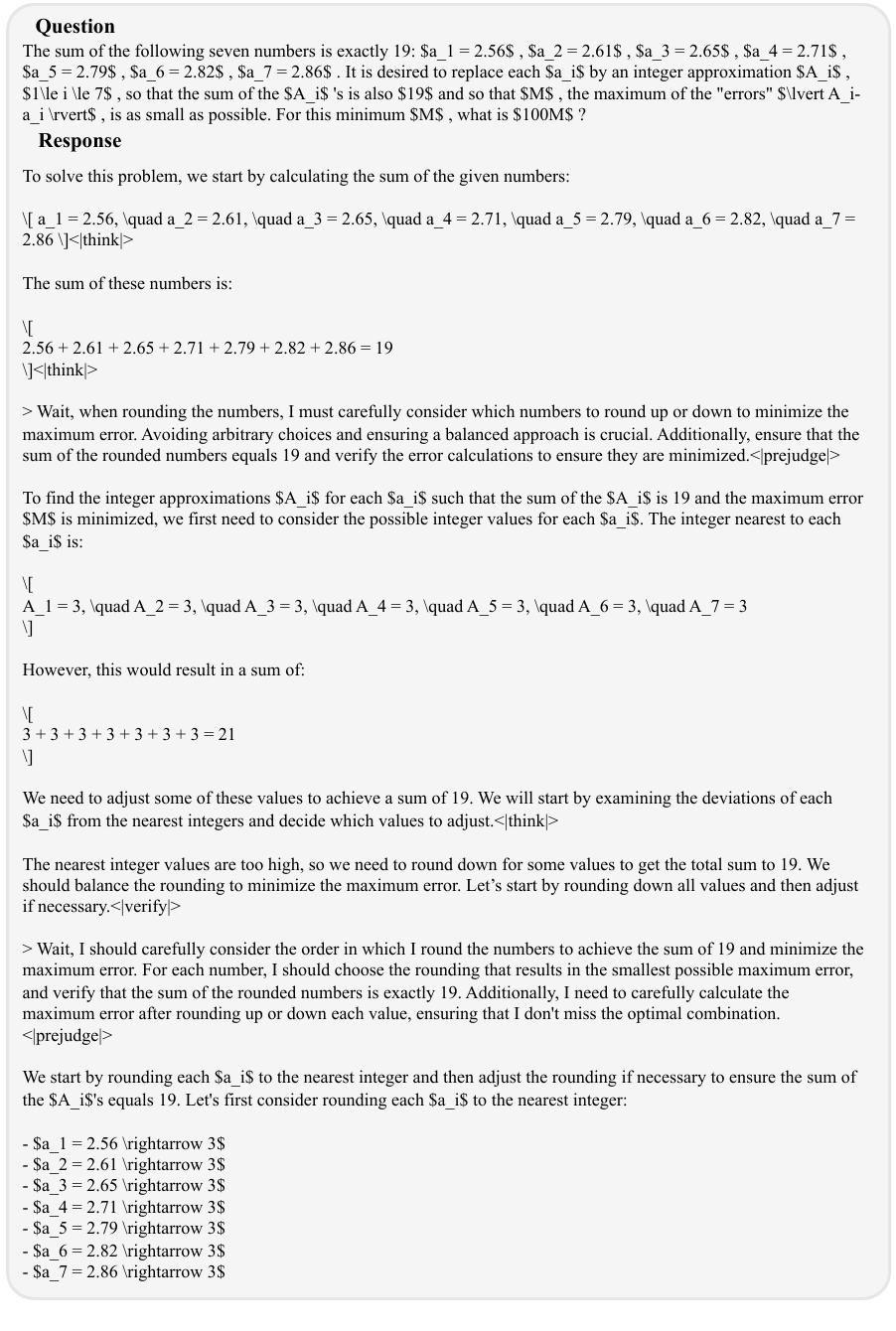} \hfill
  \caption {The case of prejudge reasoning (part 1).}
  \label{fig:case1_1}
\end{figure*}

\begin{figure*}[t]
  \includegraphics[width=\linewidth]{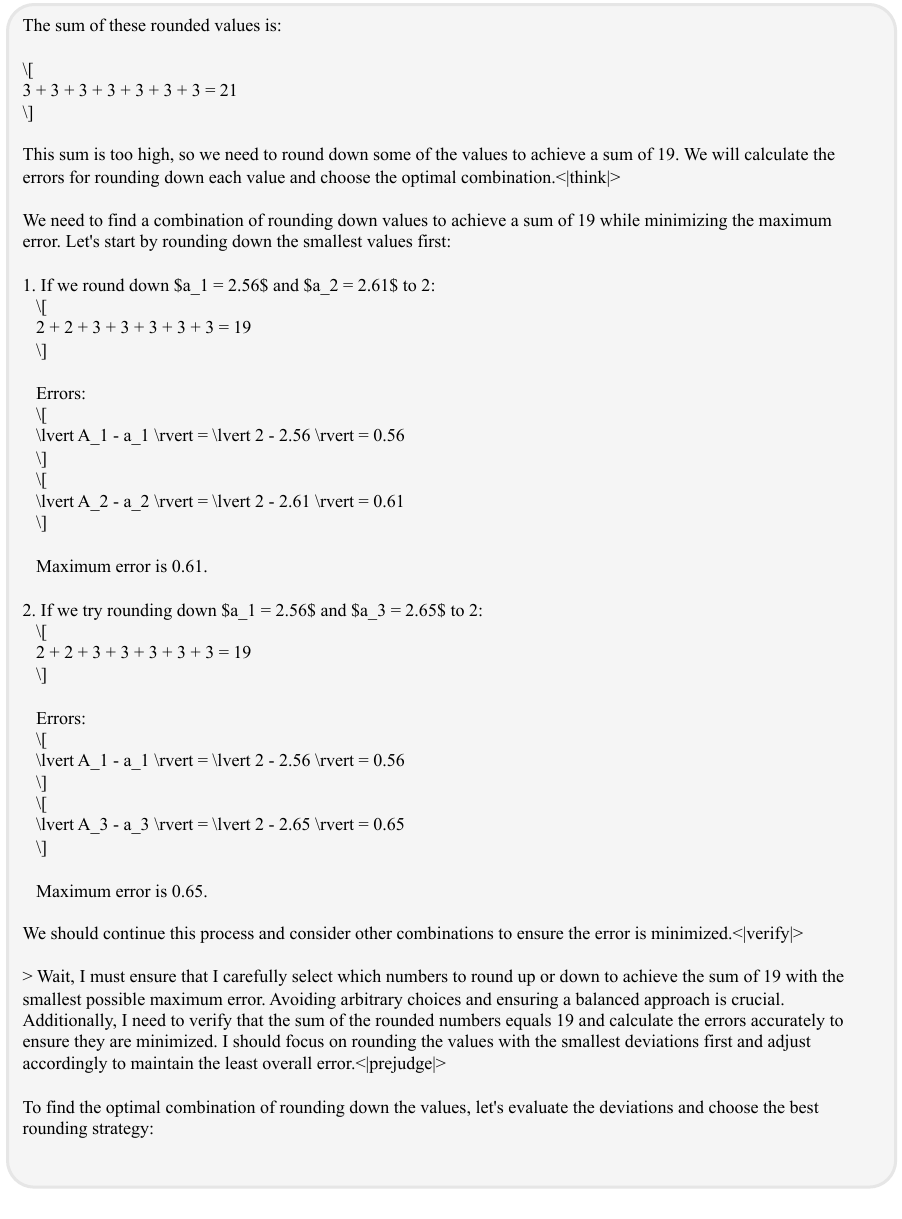} \hfill
  \caption {The case of prejudge reasoning (part 2).}
  \label{fig:case1_2}
\end{figure*}

\begin{figure*}[t]
  \includegraphics[width=\linewidth]{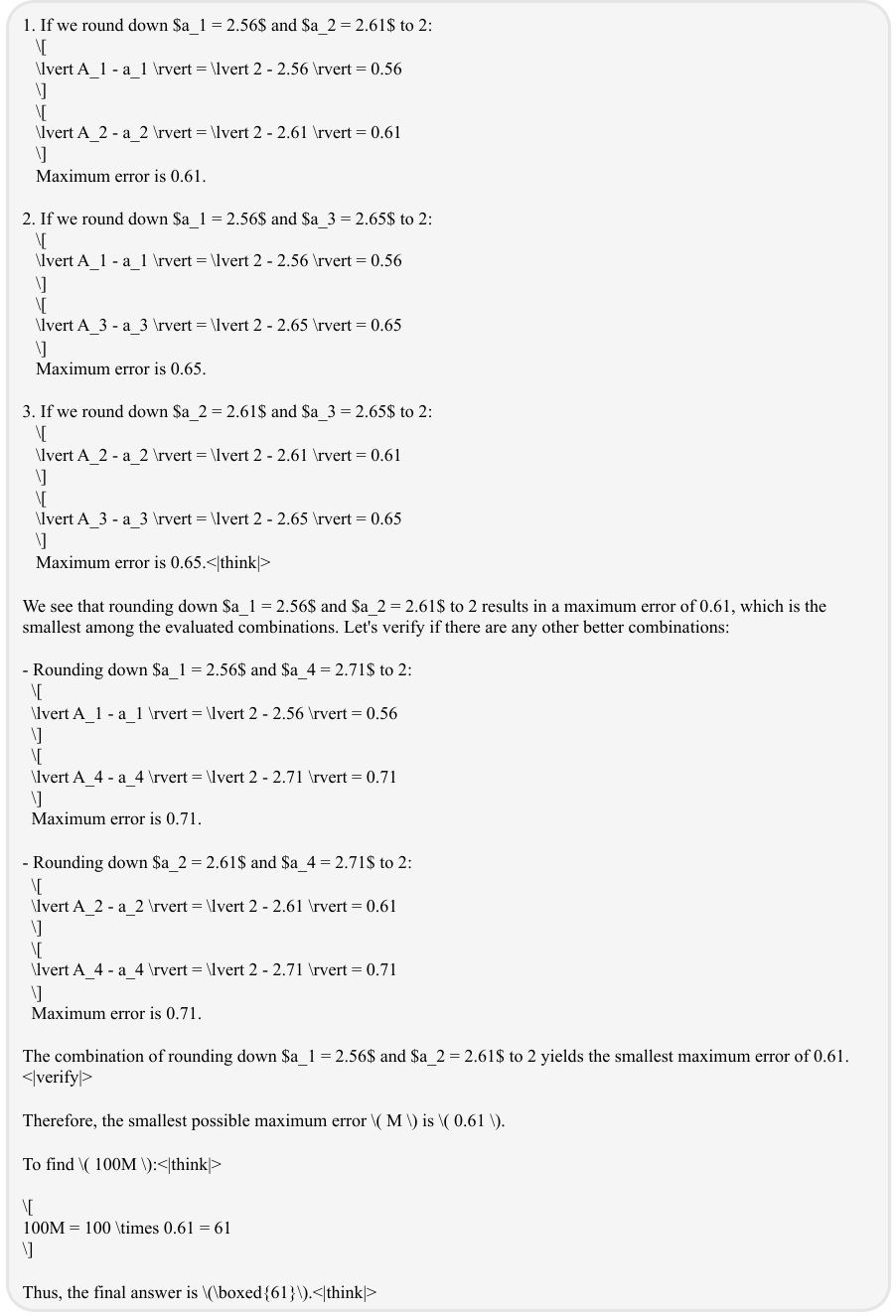} \hfill
  \caption {The case of prejudge reasoning (part 3).}
  \label{fig:case1_3}
\end{figure*}

To show how the LLM reasoning by prejudge, we provide two cases to demonstrate the rationale.
The cases are shown in Figure~\ref{fig:case1_1} to Figure~\ref{fig:case1_3}.

\end{document}